\documentclass[journal,10pt,final]{IEEEtran}

\pdfoutput=1

\usepackage{amsmath}
\usepackage{amssymb}
\usepackage{amsthm}
\usepackage{mathtools}
\usepackage[pdftex]{graphicx}
\usepackage{tabularx}
\usepackage{caption}
\usepackage{subcaption}
\usepackage{placeins}
\usepackage{algorithm}
\usepackage[noend]{algpseudocode}
\usepackage{multirow}
\usepackage{soul}
\usepackage{color}
\usepackage{bm}
\usepackage{dsfont}
\usepackage{cite}
\usepackage{hyperref}

\newcommand{\R}{\mathbb{R}}

\newcommand{\SVT}{\mathbf{SVT}}
\renewcommand{\vec}{\mathbf{vec}}
\newcommand{\OptShrink}{\mathbf{OptShrink}}
\newcommand{\soft}{\mathbf{soft}}
\newcommand{\TV}{\mathbf{TV}}
\newcommand{\WTV}{\overline{\mathbf{TV}}}
\newcommand{\TVDN}{\mathbf{TVDN}}
\newcommand{\prox}{\mathbf{prox}}
\newcommand{\sign}{\mathbf{sign}}
\newcommand{\diag}{\mathbf{diag}}
\newcommand{\ie}{i.e.,~}
\newcommand{\eg}{e.g.,~}
\newcommand{\proj}{\mathcal{P}}
\newcommand{\convas}{\overset{\textrm{a.s.}}{\longrightarrow}}
\newtheorem{theorem}{Theorem}

\begin{document}

\title{Panoramic Robust PCA for Foreground-Background Separation on Noisy, Free-Motion Camera Video}

\author{
Brian~E.~Moore\textsuperscript{*}, \IEEEmembership{Member,~IEEE},
Chen~Gao\textsuperscript{*}, \IEEEmembership{Student~Member,~IEEE}, and
Raj~Rao~Nadakuditi, \IEEEmembership{Member,~IEEE}
\thanks{Asterisk (*) denotes equal contribution.}
\thanks{This work was supported in part by the following grants: ONR grant N00014-15-1-2141, DARPA Young Faculty Award D14AP00086, and ARO MURI grants W911NF-11-1-0391 and 2015-05174-05.}
\thanks{B.~E.~Moore, C.~Gao, and R.~R.~Nadakuditi are with the Department of Electrical Engineering and Computer Science, University of Michigan, Ann Arbor, MI 48109 USA (email: brimoor@umich.edu; chengao@umich.edu; rajnrao@umich.edu).}
}

\markboth{IEEE TRANSACTIONS ON COMPUTATIONAL IMAGING}{}

\maketitle

\IEEEpeerreviewmaketitle

\begin{abstract}
This work presents a new robust PCA method for foreground-background separation on freely moving camera video with possible dense and sparse corruptions.
Our proposed method registers the frames of the corrupted video and then encodes the varying perspective arising from camera motion as missing data in a global model. This formulation allows our algorithm to produce a panoramic background component that automatically stitches together corrupted data from partially overlapping frames to reconstruct the full field of view. We model the registered video as the sum of a low-rank component that captures the background, a smooth component that captures the dynamic foreground of the scene, and a sparse component that isolates possible outliers and other sparse corruptions in the video. The low-rank portion of our model is based on a recent low-rank matrix estimator (OptShrink) that has been shown to yield superior low-rank subspace estimates in practice. To estimate the smooth foreground component of our model, we use a weighted total variation framework that enables our method to reliably decouple the true foreground of the video from sparse corruptions. We perform extensive numerical experiments on both static and moving camera video subject to a variety of dense and sparse corruptions. Our experiments demonstrate the state-of-the-art performance of our proposed method compared to existing methods both in terms of foreground and background estimation accuracy.
\end{abstract}

\begin{IEEEkeywords}
Robust PCA, foreground-background separation, total variation, computer vision, random matrix theory, machine learning.
\end{IEEEkeywords}

\section{Introduction} \label{sec:intro}

Principle component analysis (PCA) is an important method in signal processing and statistics for uncovering latent low-rank structure in high dimensional datasets. In turn, low-rank structure is an important model in computer vision because the high temporal correlation of video naturally admits a low-rank representation. Although PCA is stable in the presence of relatively small noise, it is well-known that even a few large outliers in the data can cause PCA to breakdown completely.

To mitigate the breakdown of PCA, robust PCA algorithms have recently been proposed that seek to decompose a data matrix into a low-rank component and a sparse component. Recent works \cite{candes2011robust,chandrasekaran2011rank} have established that one can exactly recover the low-rank and sparse components of a matrix $Y$ under some mild assumptions in the noiseless setting by solving a convex optimization problem of the form
\begin{equation} \label{eq:rpca:noiseless}
\begin{array}{r@{~~}l}
\displaystyle\min_{L,S} & \|L\|_{\star} + \lambda \|S\|_1 \\
\text{s.t.} & Y = L + S,
\end{array}
\end{equation}
where $\|L\|_{\star}$ is the nuclear norm (sum of singular values) of the low-rank component and $\|S\|_1$ is the elementwise $\ell_1$ norm of the sparse component. Simple alternating algorithms exist \cite{candes2011robust} for solving \eqref{eq:rpca:noiseless}, which has led to widespread adoption of robust PCA methods in practice.

Robust PCA has found many applications in computer vision problems. For example, in \cite{zhang2010tilt} a robust PCA-based method is developed to learn low-rank textures from corrupted two-dimensional (2D) images of a 3D scene. Or in \cite{peng2012rasl} robust PCA is used to align a batch of linearly correlated images in the presence of gross corruptions such as occlusions. Other applications of robust PCA in computer vision include subspace segmentation and feature extraction \cite{liu2011latent}, robust subspace clustering \cite{elhamifar2013sparse}, and object segmentation \cite{minaee2016screen,minaee2017masked}. See \cite{udell2016generalized} for an overview of popular low-rank models.

In this work, we focus on another important problem in computer vision: foreground-background separation. In particular, we are interested in robust foreground-background separation, where one decomposes a scene into a static background component and a dynamic foreground component in the presence of corruptions. Such decompositions are valuable in vision applications because the components contain useful information for subsequent processing. For example, the foreground component is useful for motion detection \cite{huang2011advanced}, object recognition \cite{tsaig2002automatic}, moving object detection \cite{bouwmans2014robust,sobral2014comprehensive} and video coding \cite{ye2015foreground}. The background component can also be useful in applications such as background subtraction \cite{elgammal2000non,piccardi2004background}, where one estimates a background model of a scene and then discriminates moving objects by subtracting the model from new frames. The paper \cite{bouwmans2014robust} provides an overview of robust PCA methods for video surveillance applications.

\subsection{Background} \label{subsec:background}
There has been substantial work on foreground-background separation. For example, in \cite{elgammal2000non} the authors propose a non-parametric model for background subtraction, and a probabilistic background model for tracking applications is developed in \cite{rittscher2000probabilistic}. Alternatively, supervised approaches like GMM \cite{stauffer1999adaptive} learn a model of the background from labeled training data. Other lines of research have focused on performing background subtraction when the background is known to contain dynamic elements. Examples include a motion-based model \cite{mittal2004motion} that utilizes adaptive kernel density estimation and an online autoregressive model \cite{ramesh2003background} for modeling and subtracting dynamic backgrounds from scenes. In \cite{zhong2003segmenting} a robust Kalman filter-based approach is developed to segment foreground objects from dynamic textured backgrounds. A subspace learning method was proposed in \cite{minaee2017subspace} that learns a signal model that can remove structured outliers (foreground) from background images. The paper \cite{piccardi2004background} surveys the popular background subtraction methods in the literature.

More recently, robust PCA methods have been proposed \cite{candes2011robust,chandrasekaran2011rank,zhou2010stable,guyon2012foreground,zhou2013shifted,netrapalli2014nonconvex} that decompose video into a low-rank component containing the background and a spatially sparse component that captures the foreground of the scene. Typically the original robust PCA problem \eqref{eq:rpca:noiseless} is extended to the noisy case by relaxing the equality constraint to an inequality constraint, \eg as in \cite{xu2010robust}, or adding a data fidelity term and solving an unconstrained problem. One can easily extend the robust PCA model to the general inverse problem setting by introducing a linear operator in the data fidelity term. Of particular interest in this work is the model proposed in \cite{otazo2015low}, which we refer to as the RPCA method. Many other extensions to the robust PCA framework have been proposed and applied to perform foreground-background separation. The Fast PCP method \cite{rodriguez2013fast} models the background with an explicit rank constraint and proposes an efficient implementation of principal component pursuit to compute the decomposition. Other methods that propose changes to the background model/penalty include the noncvxRPCA method \cite{kang2015robust}, which uses a non-convex surrogate for the rank penalty to improve the accuracy of the estimated background, and the ROSL method \cite{shu2014robust}, which proposes a novel rank measure based on a group sparsity assumption on the coefficients of the background in an orthonormal subspace. Another method called MAMR \cite{ye2015foreground} uses a dense motion field to compute a weight matrix in the objective function that represents the likelihood that each pixel is in the background. See \cite{bouwmans2016handbook,vaswani2018robust} for a comprehensive survey of robust PCA methods for image and video processing applications.

Although standard sparsity-based foreground models are effective in the noiseless scenario, they are unable to distinguish foreground from sparse corruptions. In this context, models employing total variation (TV) have been proposed to model the spatial continuity of the foreground of a scene \cite{guyon2012foreground,guyon2012foreground2}. Recently, the TVRPCA method \cite{cao2016total} was proposed to separate dynamic background from moving objects using TV-based regularization, which demonstrates that TV-based models can effectively distinguish foreground from sparse corruptions.

Another important class of foreground-background separation models are those that can handle dynamic scenes arising, \eg from moving camera video. In such cases, the background of the raw video may not be low-rank, so care is required to map the problem to an appropriate model that recovers low-rank structure. One approach for moving camera video is to adopt an online learning framework where batches of frames are sequentially processed and the foreground-background model is sequentially updated based on the latest batch. A popular approach is GRASTA \cite{he2011online}, which models the background as a subspace on the Grassmannian manifold and develops an iterative algorithm for tracking the low-rank subspace. The Prac-REProCS method \cite{guo2014online} proposes an online method based on alternating projection, sparse recovery, and low-rank subspace updates. The Layering Denoising method \cite{guo2016video} is a recent extension of REProCS that performs a video denoising step on the estimated background and foreground components at each iteration. Finally, the recent IncPCP-PTI method \cite{chau2017panning} proposes an extension of principal component pursuit that iteratively aligns the estimated background component to the current reference frame. See \cite{yazdi2018new} for a recent survey of object detection methods designed for the moving camera setting.

Such online methods typically require the subspaces to evolve relatively slowly over time in order to accurately track them, which precludes their application on many practical videos with significant camera motion. Other methods use parametric models to estimate the transformations that describe the motion in the scene. The RASL method \cite{peng2012rasl} proposes a robust PCA model that iteratively estimates the decomposition along with the parameters of an affine transformation model, but this approach is computationally expensive in practice. An efficient implementation of this idea was proposed in \cite{ebadi2016approximated}, but, in both cases, the models consider only the intersection (common view) of the scene. A popular method is DECOLOR \cite{zhou2013moving}, which employs $\ell_0$-based regularization and a Markov random field model to jointly estimate the dynamic background and the foreground support.

Recently deep learning-based methods for background subtraction have also shown promise. For example, the method from \cite{braham2016deep} models the background of a video as an evolving weightless neural network that learns the distribution of pixel color in the video. Alternatively, the authors of \cite{de2017background} propose a background subtraction algorithm that uses convolutional neural networks trained on spatial video features to subtract the background patch-wise from a video based on an a priori model of the scene.
Another supervised learning-based approach is to employ a semantic segmentation model trained on a labeled dataset containing objects of interest to directly produce foreground masks for each frame of a video. Deep convolutional models in particular have shown great promise for semantic segmentation, and examples of recent models include Mask R-CNN \cite{he2017mask}, the MIT Scene Parsing benchmark \cite{zhou2017scene}, DeepLab \cite{chen2018deeplab}, and Deformable DeepLab \cite{dai2017deformable}. Such supervised methods are appealing in situations where the classes of foreground objects are known in advance, but unsupervised techniques like robust PCA are applicable to datasets with arbitrary semantic content. While deep learning-based methods have shown promise in the noiseless regime, they are generally not designed to perform foreground-background separation in the presence of corruptions, which is our focus in this work.

\subsection{Contributions} \label{subsec:contrib}
In this paper, we propose a robust foreground-background separation method based on the robust PCA framework that can decompose a corrupted video with freely moving camera into a panoramic low-rank background component and a smooth foreground component. Our algorithm proceeds by registering the frames of the raw video to a common reference perspective and then minimizing a modified robust PCA cost that accounts for the unobserved data resulting from the partially overlapping views of the registered frames.

Our proposed method advances the state-of-the-art in several key aspects. First, our method produces a panoramic background component that spans the entire field of view, whereas existing parametric models typically only estimate the subspace spanning the intersection of the views. This panoramic property is useful because it allows one to produce a denoised version of the entire moving camera video. Our background model also employs an improved low-matrix estimator (OptShrink) \cite{nadakuditi2014optshrink} that has been shown to yield superior subspace estimates in practice compared to singular value thresholding-based approaches \cite{moore2014improved,ravishankar2017lassi}. Our method also separates the dynamic foreground of a scene from sparse corruptions using TV regularization. Most existing foreground-background separation methods are not designed to disentangle the foreground from additional (sparse or dense) corruptions, so this capability is significant. Our numerical experiments indicate that our formulation produces more accurate foreground estimates compared to existing TV-based methods (which do not handle moving cameras). We account for the deforming view in the registered frames by considering a weighted total variation penalty that omits differences involving unobserved pixels, and we propose an efficient algorithm for minimizing this objective. Although the applicability of our proposed method to video with significant camera motion is interesting, we also note that our proposed OptShrink-based background model and our particular formulation of TV-based foreground regularization yield state-of-the-art results on the important and well-studied problem of robust foreground-background separation on \emph{static camera video}, which is of independent interest in many applications.

A similar panoramic frame registration procedure was proposed in the video background subtraction method from \cite{mansour2014video}. However, our method extends this work in several key respects. In particular, we are focused on the noisy/corrupted video setting, whereas the method from \cite{mansour2014video} is designed for noiseless video. We achieve this robustness by adopting the aforementioned OptShrink-based low-rank estimator in lieu of a matrix factorization-based model, which allows our method to produce robust background estimates. Our inclusion of a TV-regularized component in addition to an $\ell_1$-sparsity-regularized component also allows our method to denoise and disentangle the foreground from possible sparse corruptions, whereas the method from \cite{mansour2014video} includes only one sparse component, which cannot distinguish foreground from corruptions.

A short version of this work was recently presented elsewhere \cite{gao2017augmented}. Here, we build substantially on this work by performing extensive numerical experiments comparing our proposed method to state-of-the-art methods in both the static and moving camera settings. In our numerical experiments, we consider both dense noise and sparse outliers. We also improve the computational efficiency of the total variation-related components of our proposed method.

\subsection{Organization} \label{subsec:organization}
The paper is organized as follows. In Section~\ref{sec:registration}, we describe our video registration strategy. Section~\ref{sec:form} formulates our proposed augmented robust PCA model, and we present our algorithm for solving it in Section~\ref{sec:algo}. In Section~\ref{sec:experiments}, we provide extensive numerical experiments that demonstrate the state-of-the-art performance of our method compared to existing methods on both static camera and moving camera videos under a variety of corruption models. Finally, Section~\ref{sec:conclusion} concludes and discusses opportunities for future work.

\begin{figure*}[t!]
\centering
\centerline{\includegraphics[width=\textwidth]{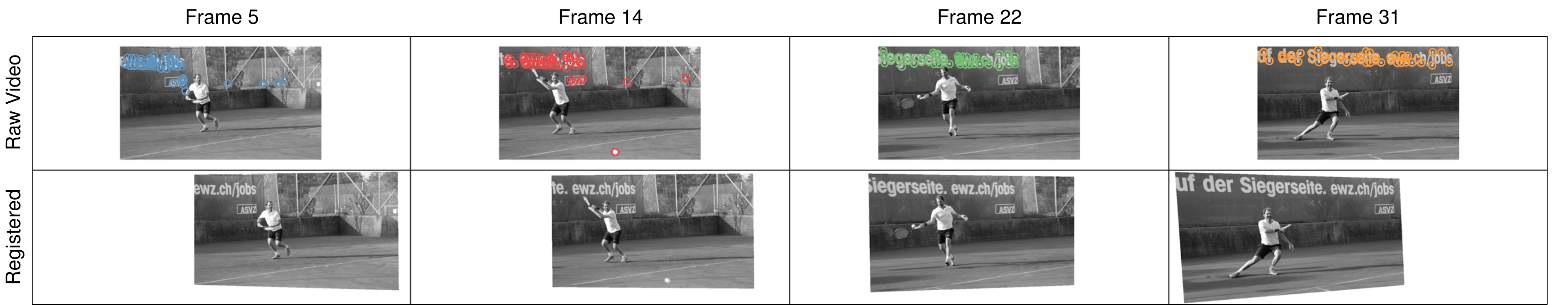}}
\caption{The video registration process. The top row depicts raw video frames $F_k$ with SURF features annotated. The bottom row depicts the corresponding registered frames $\widetilde{F}_k$ computed via \eqref{eq:registration}. The $k$th column of the mask matrix $M \in \{0,1\}^{mn \times p}$ encodes the support of $\widetilde{F}_k$ within the aggregate view; \ie $M_{ik} = 0$ for unobserved pixels, which are represented by white space in the registered frames above.}
\label{fig:frames}
\vspace{-0.3cm}
\end{figure*}

\section{Video Registration} \label{sec:registration}
The vast majority of video data gathered today is captured by moving (\eg handheld) cameras. To process this data in a robust PCA framework, our approach is to register the raw video---\ie map the frames to a common reference perspective---and then jointly process the registered data. In this work, we adopt the standard perspective projection model \cite{forsyth2002computer}, which relates different views of a scene via homographic transformations.

\subsection{Registering two frames} \label{subsec:register:two}
Consider a point $(x,y)$ in a frame that is known to correspond to a point $(\widetilde{x},\widetilde{y})$ in another frame. Under a planar surface model, one can relate the points via the projective transformation
\begin{equation} \label{eq:projective}
\kappa\widetilde{p} = H^Tp,
\end{equation}
where $\widetilde{p} = [\widetilde{x}, \widetilde{y}, 1]^T$, $p = [x, y, 1]^T$, $\kappa \neq 0$ is an arbitrary scaling constant, and $H \in \mathbb{R}^{3 \times 3}$ with $H_{33} = 1$ is the unknown projective transformation matrix. Given $d > 3$ correspondences $\{(x_i, y_i) \mapsto (\widetilde{x}_i, \widetilde{y}_i)\}_{i=1}^d$, one can estimate $H$ in a least squares sense by minimizing \cite{forsyth2002computer}
\begin{equation} \label{eq:projective_cost}
\min_h \|Ah\|^2 \text{~~s.t.~~} h_9 = 1,
\end{equation}
where $h = \vec(H)$ is the vectorized version of $h$ formed by stacking the columns of $H$ into a vector, $A^T = \begin{bmatrix} A_1^T, \ldots , A_d^T \end{bmatrix}$, and
\begin{equation}
A_i = \begin{bmatrix}
0 & p_i^T & -\widetilde{y}_i p_i^T\\
p_i^T & 0 & -\widetilde{x}_i p_i^T
\end{bmatrix} \in \mathbb{R}^{2 \times 9}.
\end{equation}
The solution to \eqref{eq:projective_cost} is the smallest right singular vector of $A$, scaled so that the last element is 1.

To estimate $H$ in practice, one must also generate correspondences $(x_i, y_i) \mapsto (\widetilde{x}_i, \widetilde{y}_i)$ between points in the frames. In this work, we adopt the standard procedure \cite{forsyth2002computer} of computing SURF features \cite{bay2006surf} for each frame and then using the RANSAC \cite{fischler1981random} algorithm to find a robust subset of the correspondences that produce a solution $\hat{H}$ to \eqref{eq:projective_cost} with small cost. Importantly, this robust approach can generate accurate transformations in the presence of corruptions in the raw video.


\subsection{Registering a video} \label{subsec:register:video}
One can readily extend the two-frame registration procedure from Section~\ref{subsec:register:two} to a video by iteratively constructing homographies $H_k := H_{k \mapsto k+1}$ between frames $k$ and $k + 1$ of the video and then composing the homographies to map all $p$ frames to an anchor frame $\widetilde{k} \in \{1,\ldots,p\}$. Here, we choose the middle frame $\widetilde{k} = \lfloor p/2 \rfloor$. Consecutive frames of a video are highly correlated, so the homographies $H_k$ can be computed with high accuracy.

Let $F_1, \dots, F_p \in \mathbb{R}^{a \times b}$ denote the frames of a moving camera video, and denote by $\mathcal{H}_k \coloneqq \mathcal{H}_{k \mapsto k+1}$ the linear transformation that applies the projective transformation \eqref{eq:projective} defined by $H_k$ to each pixel of $F_k$. One can register the frames with respect to anchor frame $\widetilde{k}$ by computing
\begin{equation} \label{eq:registration}
\widetilde{F}_k =
\begin{cases}
(\mathcal{H}_{\widetilde{k} - 1} \circ  \mathcal{H}_{\widetilde{k} - 2} \circ \dots \circ \mathcal{H}_k)(F_k) & k < \widetilde{k}, \\
\hphantom{(} F_k & k = \widetilde{k}, \\
(\mathcal{H}_{\widetilde{k}}^{-1} \circ  \mathcal{H}_{\widetilde{k} + 1}^{-1} \circ \dots \circ \mathcal{H}_{k - 1}^{-1})(F_k) & k > \widetilde{k},
\end{cases}
\end{equation}
for each $k = 1,\ldots,p$. The above procedure yields $\widetilde{F}_1,\ldots,\widetilde{F}_p \in \mathbb{R}^{m \times n}$, a collection of registered frames in a common perspective, where $m$ and $n$ are the height and width of the region defined by the union of the registered frame extents. See Figure~\ref{fig:frames} for a graphical depiction of this procedure applied to a moving camera video.

The registered frames $\widetilde{F}_k$ form a static camera video in the sense that a given coordinate $(\widetilde{F}_k)_{ij}$ now corresponds to the same spatial location for each frame $k$. If the composite projective transformation mapping $F_k$ to $\widetilde{F}_k$ is not the identity transformation, the matrix $\widetilde{F}_k$ will contain some pixels that correspond to locations outside the view of the original frame $F_k$. Without loss of generality, we set such unobserved pixels to zero in $\widetilde{F}_k$.

\section{Problem Formulation} \label{sec:form}
In this section, we describe our proposed robust PCA framework for panoramic foreground-background separation. We first describe our model, discuss our treatment of total variation for moving camera video, and then we present our problem formulation. We discuss our algorithm for solving the proposed problem in Section~\ref{sec:algo}.

\subsection{Data Model} \label{subsec:model}
Given the registered frames $\widetilde{F}_1,\ldots,\widetilde{F}_p \in \mathbb{R}^{m \times n}$ of a moving camera video computed as in Section~\ref{sec:registration}, we construct the data matrix $Y \in \mathbb{R}^{mn \times p}$
\begin{equation} \label{eq:Ymatrix}
Y = \begin{bmatrix}
\vec(\widetilde{F}_1) \dots \vec(\widetilde{F}_p)
\end{bmatrix}
\end{equation}
whose columns are the vectorized registered frames. Associated with $Y$, we also define the mask matrix $M \in \{0, 1\}^{mn \times p}$ whose columns encode the support of the registered frames in the aggregate view. See Figure~\ref{fig:frames} for a graphical depiction.\footnote{When processing video that is known (or modeled) to have a static camera, one can omit the video registration step and directly construct the data matrix $Y$ by vectorizing each frame of the raw video. In this case, the corresponding mask matrix $M$ is the all-ones matrix.}

The representation \eqref{eq:Ymatrix} is useful because each row of $Y$ corresponds to a fixed point in space, so we have effectively reduced the moving camera foreground-background separation problem to a static camera problem with incomplete observations (corresponding to the zeros in $M$). Thus, with suitable modifications to account for the missing data, we can readily apply ideas from standard static camera foreground-background separation. In particular, our approach is to model the observed data $Y$ with a decomposition of the form
\begin{equation} \label{eq:model}
\proj_{M}(Y) = \proj_{M}(L + S + E + N),
\end{equation}
where $\proj_{M}$ denotes the orthogonal projection onto $M$:
\begin{equation} \label{eq:proj:matrix}
[\proj_{M}(X)]_{ij} = \begin{cases}
X_{ij} & M_{ij} = 1 \\
0 & M_{ij} = 0.
\end{cases}
\end{equation}
In \eqref{eq:model}, $L$ represents the (registered) background of the video, and $S$ represents the foreground. Furthermore, the matrix $E$ captures possible sparse corruptions in the video, and $N$ captures possible dense corruptions. Note that the projection operators in \eqref{eq:model} exclude unobserved pixels from our model, so we are not attempting to impute the unobserved pixels of the scene; rather we are expressing the moving camera video as a ``static" space-time matrix where each row corresponds to a fixed point in space.

Since our data is registered, the background will have high temporal correlation and thus can be well-modeled as a low-rank matrix \cite{cao2016total}. In the standard robust PCA model \cite{candes2011robust}, the foreground component $S$ is modeled as a sparse matrix. However, we are interested in reliably estimating the foreground in the presence of sparse corruptions, so a sparse model for $S$ will be indistinguishable from the sparse corruptions. Instead, we model $S$ as a smoothly-varying matrix and, motivated by the recent TVRPCA method \cite{cao2016total}, use a total variation-based regularization framework to estimate $S$. In the moving camera setting, we will consider a weighted total variation penalty that avoids penalizing first differences involving unobserved pixels. Motivated by the vast compressed sensing literature \cite{donoho2006compressed,candes2006robust}, we model $E$ as a sparse matrix and employ $\ell_1$-based regularization to estimate it. Finally, we model $N$ as a dense noise matrix, and we estimate it by imposing the familiar least-squares-based regularization. We explicitly describe the optimization problem that we employ to learn the model \eqref{eq:model} in Section~\ref{subsec:problem}.


\subsection{Weighted Total Variation} \label{subsec:tv}
Total variation regularization is a ubiquitous method in image and video processing for reconstructing signals corrupted by noise \cite{rudin1992nonlinear,chambolle2004algorithm,chan2011augmented}. In particular, in this work, given a matrix $X \in \R^{mn \times p}$ whose columns contain the vectorized $m \times n$ frames of a video, we consider the \emph{weighted anisotropic TV} of $X$:

\begin{equation} \label{eq:tv}
\WTV(X) = \sum_{ijk} \left( \begin{split}
w^x_{ijk}|&x_{i+1jk} - x_{ijk}| + w^y_{ijk}|x_{ij+1k} - x_{ijk}| \\
& \quad + w^z_{ijk}|x_{ijk+1} - x_{ijk}|
\end{split} \right)
\end{equation}

Here, we use a slight abuse of notation by implicitly referencing the vectorized video $x = \vec(X) \in \R^{mnp}$ in the definition and using $x_{ijk}$ to denote the pixel $(i,j)$ from frame $k$ of the video---\ie the $(i + m(j - 1),~k)$ entry of $X$. In \eqref{eq:tv}, $w^x_{ijk}$, $w^y_{ijk}$, and $w^z_{ijk}$ are fixed $\{0,1\}$ weights that omit first differences involving the unobserved pixels that lie outside the extent of the registered frames. These weights can be readily computed from the mask matrix $M$ (see Figure~\ref{fig:frames}). We omit the summation indices in \eqref{eq:tv} for brevity, but it should be understood that we are not considering the first differences corresponding to circular boundary conditions in our model (\eg $x_{1jk} - x_{mjk}$).

Smoothly varying objects with few sharp edges will have low TV, so \eqref{eq:tv} is a good model for the foreground of a video \cite{rudin1992nonlinear,chambolle2004algorithm,chan2011augmented}. Conversely, sparse corruptions will have very high TV, so it is reasonable to expect that \eqref{eq:tv} will be able to distinguish the foreground from sparse corruptions.

Subsequently, we will refer to \eqref{eq:tv} as three-dimensional (3D) TV because it penalizes both the spatial first differences between neighboring pixels in a given frame and the temporal differences between a given pixel in consecutive frames. Such a model may be appropriate for datasets with high temporal correlation, \eg due to a slowly moving camera. However, in other cases, it may be preferable to omit the temporal differences from \eqref{eq:tv} by setting $w^z_{ijk} = 0$. We refer to this latter model as 2D TV.

\subsubsection{Matrix-Vector Representation} \label{subsubsec:tv:vec}
When describing our proposed algorithm in Section~\ref{subsec:problem}, it will be convenient for us to express the TV penalty \eqref{eq:tv} using matrix-vector operations.

In the 1D case, one can compute the first differences of $z \in \mathbb{R}^n$ with the matrix-vector product $D_n z$, where $D_n \in \mathbb{R}^{n \times n}$ is the circulant first differences matrix
\begin{equation} \label{eq:Dn}
D_n = \begin{bmatrix}
-1 &  1 &        &        &    \\
   & -1 &      1 &        &    \\
   &    & \ddots & \ddots &    \\
   &    &        &     -1 &  1 \\
 1 &    &        &        & -1
\end{bmatrix}.
\end{equation}
Note that we include the first difference $[D_n z]_n = z_1 - z_n$ corresponding to circular boundary conditions in this computation, although we omit these circular differences from our TV penalty \eqref{eq:tv}. We do this because we will later leverage the fact that $D_n$ is a circulant matrix. Using this notation, we can write the 1D TV penalty as $\WTV(z) = \|W D_n z\|_1$, where $W$ is the diagonal matrix with $W_{kk} = 1$ for $k < n$ and $W_{nn} = 0$, which omits the circular boundary difference. In general, one can omit other first differences by setting the corresponding diagonal entry of $W$ to zero.

In the 3D case when $x \in \R^{mnp}$, one can compute the first differences along each dimension of the vectorized $m \times n \times p$ tensor by computing the matrix-vector product $Cx$, where $C \in \R^{3mnp \times mnp}$ is the matrix
\begin{equation} \label{eq:cmat}
C = \begin{bmatrix}
I_p \otimes I_n \otimes D_m \\
I_p \otimes D_n \otimes I_m \\
D_p \otimes I_n \otimes I_m
\end{bmatrix}.
\end{equation}
In \eqref{eq:cmat}, $I_n$ is the $n \times n$ identity matrix and $\otimes$ denotes the Kronecker product. Again, we have included the first differences corresponding to circular boundary conditions for mathematical convenience so that $C$ is constructed from Kronecker products of circulant matrices. Using this definition, we can write
\begin{equation} \label{eq:tv:vec}
\WTV(X) = \|WCx\|_1,
\end{equation}
where $W$ is the diagonal $\{0,1\}$ matrix that omits first differences involving unobserved pixels and those corresponding to circular boundary conditions. Concretely, one has
\begin{equation} \label{eq:tv:wmat}
W = \diag(\vec(w^x), \vec(w^y), \vec(w^z)),
\end{equation}
where $w^x$, $w^y$, and $w^z$ are the $m \times n \times p$ tensors containing the weights from \eqref{eq:tv} and we set $w^x_{mjk} = w^y_{ink} = w^z_{ijp} = 0$ for all $ijk$ to omit the circular boundaries. Here, $\vec(\cdot)$ converts an $m \times n \times p$ tensor into a vector by stacking the columns of each frame into length-$mn$ vectors and then stacking these vectors to form a single length-$mnp$ vector, and $\diag(\cdots)$ constructs a diagonal matrix from the vector formed by concatenating its vector arguments into a single vector. We will rely on the equivalent representation \eqref{eq:tv:vec} of \eqref{eq:tv} when presenting our proposed algorithm in Section~\ref{subsec:problem}.

\subsection{Proposed Optimization Problem} \label{subsec:problem}
To learn a decomposition of the form \eqref{eq:model}, we propose to solve the augmented robust PCA problem
\begin{equation} \label{eq:cost:full}
\begin{array}{rl}
\displaystyle\min_{L,S,E,N} & \lambda_L \|L\|_{\star} + \lambda_S \WTV(S) + \lambda_E \|E\|_1 + \tfrac{1}{2}\|N\|_F^2 \\[8pt]
\text{s.t.} & \proj_{M}(Y) = \proj_{M}(L + S + E + N).
\end{array}
\end{equation}
Equivalently, one can eliminate matrix $N$ in \eqref{eq:cost:full} and instead consider the unconstrained problem
\begin{equation} \label{eq:cost}
\begin{array}{rl}
\displaystyle\min_{L,S,E} & \tfrac{1}{2}\|\proj_{M}(Y - L - S - E)\|_F^2 ~+ \\
& \quad \qquad \lambda_L \|L\|_{\star} + \lambda_{S} \WTV(S) + \lambda_{E}\|E\|_1.
\end{array}
\end{equation}
Here, $\WTV(\cdot)$ is the weighted TV penalty defined in \eqref{eq:tv} and the parameters $\lambda_L,\lambda_S,\lambda_E \geq 0$ are regularization parameters that control the relative contribution of each term to the overall cost. It is well-known that each term in \eqref{eq:cost} is a convex function, so \eqref{eq:cost} is a convex problem.

As discussed in Section~\ref{subsec:model}, the $L$ component of our model represents the background of the video, which we model as a low-rank matrix and thus regularize with the nuclear norm. The $S$ component represents the foreground, which we model as a smoothly-varying matrix with sharp edges and regularize with the weighted TV penalty. The $E$ component represents sparse corruptions, which we model as a sparse matrix and regularize with the familiar $\ell_1$ penalty. The first term in \eqref{eq:cost} is a data fidelity term that forces the decomposition $L+S+E$ to approximately agree with the data $Y$ at the observed pixel locations encoded by the mask matrix $M$. The choice of $\left\|\cdot\right\|_F^2$ for the data fidelity term captures residual dense corruptions in the data.

Note that in the moving camera scenario where $M$ is not the all-ones matrix, the entries of the $mn \times p$ matrices $L$, $S$, and $E$ corresponding to entries of $M$ such that $M_{ij} = 0$ represent unobserved pixels in the scene. Moreover, careful inspection of \eqref{eq:cost} and \eqref{eq:tv} shows that the unobserved entries $\proj^{\perp}_M(S)$ are completely omitted from the cost and thus, without loss of generality, we can set the unobserved entries of $S$ to zero. In addition, the unobserved entries of the sparse component $E$ only appear in the regularization term $\|E\|_1$, so any optimal value $\hat{E}$ must satisfy $\proj^{\perp}_M(\hat{E}) = 0$. The situation is different for $L$ because the low-rank model $\|L\|_{\star}$ allows us to impute partially unobserved background pixels in the frames (indeed, this is why the low-rank assumption is appropriate for background estimation); however, if $M$ contains any all-zero rows corresponding to pixels that are not observed in any frame, then one can show that the corresponding entries of $L$ can be set to zero.\footnote{In practice, we initialize $L = \proj_M(Y)$ and $S = E = 0$, which causes any unobserved rows of $L$ and entries of $E$ and $S$ to remain zero. In fact, for efficiency, we omit these arbitrary values from our computations.} See Figure~\ref{fig:frames} for intuition.

Our proposed problem \eqref{eq:cost} differs from the recent RPCA \cite{candes2011robust}, TVRPCA \cite{cao2016total}, and DECOLOR \cite{zhou2013moving} methods in several key ways. First, in the moving camera setting, our frame registration and masking strategy allows us to reconstruct the full field of view of the scene, while DECOLOR only estimates the overlapping (intersection) view. Second, we regularize the foreground component of our model using TV alone, while the TVRPCA method includes both $\ell_1$ and TV-based regularization on its foreground model, which is overly restrictive because the foreground need not be spatially sparse. Finally, our model improves on the standard RPCA model by including the TV-regularized component to disentangle the foreground $S$ from possible sparse corruptions, which are isolated in the $E$ component.

\section{Algorithm and Properties} \label{sec:algo}
In this section we derive our algorithm for solving \eqref{eq:cost}, present an important modification to the low-rank update, and discuss the properties of our algorithm.

\subsection{Proximal Gradient Updates} \label{subsec:updates}
We use the proximal gradient method \cite{parikh2014proximal} to minimize \eqref{eq:cost}. The proximal gradient method is an iterative algorithm for solving problems of the form $f(X) + g(X)$, where $f$ is convex and differentiable and $g$ is convex and has an easily computable proximal operator
\begin{equation} \label{eq:prox:defn}
\prox_g(Y) := \arg \min_{X} ~ \tfrac{1}{2}\|Y - X\|_F^2 + g(X).
\end{equation}
The proximal gradient method prescribes updates of the form
\begin{equation} \label{eq:prox:grad}
X^{k+1} = \prox_{\tau^k g}(X^k - \tau^k \nabla f(X^k)),
\end{equation}
where $\nabla f$ denotes the gradient of $f$ and $\tau^k > 0$ is a chosen step size. It is known \cite{parikh2014proximal} that the proximal gradient method converges when a constant step size $\tau^k = \tau < 2/L_{\nabla f}$ is used, where $L_{\nabla f}$ is the Lipschitz constant for $\nabla f$. In fact, the iterates $X^{k}$ will monotonically decrease the cost when a constant step size $\tau \leq 1 / L_{\nabla f}$ is used \cite{parikh2014proximal}.

To map \eqref{eq:cost} into a suitable form for proximal gradient, we identify $f(L,S,E) = \frac{1}{2}\|\proj_{M}(Y-L-S-E)\|_F^2$ and $g(L,S,E) = \lambda_L\|L\|_{\star} + \lambda_S\WTV(S) + \lambda_E\|E\|_1$, which we regard as functions of the single variable $X = [L~S~E]$. Under these definitions, a simple computation shows that $\nabla f = [\nabla f_L~\nabla f_S~\nabla f_E]$, where $\nabla f_L(L,S,E) = \nabla f_S(L,S,E) = \nabla f_E(L,S,E) = \proj_M(L+S+E-Y)$. Since $g$ is the sum of three functions, its proximal operator \eqref{eq:prox:defn} can be computed separately for each component. Thus our proximal update scheme for \eqref{eq:cost} can be written as
\begin{equation} \label{eq:prox:updates:1}
\begin{aligned}
U^{k+1} &= \proj_{M}(L^{k} + S^{k} + E^{k} - Y) \\[4pt]
L^{k+1} &= \prox_{\tau^k \lambda_L \|\cdot\|_{\star}} (L^k - \tau^k U^{k+1}) \\[4pt]
S^{k+1} &= \prox_{\tau^k \lambda_S \WTV} (S^k - \tau^k U^{k+1}) \\[4pt]
E^{k+1} &= \prox_{\tau^k \lambda_E \|\cdot\|_1} (E^k - \tau^k U^{k+1}),
\end{aligned}
\end{equation}
where we have introduced the auxiliary variable $U$ for notational convenience. It is straightforward to show that $L_{\nabla f} = 3$, so a constant step size $\tau < 2/3$ suffices to guarantee convergence.

The proximal operators for the $L$ and $E$ updates in \eqref{eq:prox:updates:1} have simple, closed-form solutions. Indeed, it is well-known that the solution to the nuclear-norm-regularized problem
\begin{equation} \label{eq:prox:nuclear}
\arg \min_L ~ \tfrac{1}{2}\|Z - L\|_F^2 + \lambda\|L\|_{\star}
\end{equation}
is given by the singular value thresholding operator \cite{candes2011robust,cai2010singular}
\begin{equation} \label{eq:svt}
\SVT_{\lambda}(Z) = \sum_i (\sigma_i - \lambda)_{+} u_i v_i^T,
\end{equation}
where $Z = \sum_i \sigma_i u_i v_i^T$ is the singular value decomposition (SVD) of $Z$, and $(\cdot)_{+} = \max(\cdot,0)$. The solution to the $\ell_1$-regularized problem
\begin{equation}
\arg \min_E ~ \tfrac{1}{2}\|Z - E\|_F^2 + \lambda\|E\|_{1}
\end{equation}
is given by the elementwise soft thresholding operator \cite{candes2011robust}
\begin{equation} \label{eq:soft}
\soft_{\lambda}(z) = \sign(z)(|z| - \lambda)_+.
\end{equation}

The proximal operator for the weighted TV penalty \eqref{eq:tv} does not have a closed-form solution in general,\footnote{There is a closed-form solution in the special case of static camera video when circular boundary conditions are allowed in the TV penalty. In this case, the $W$ matrix in \eqref{eq:tv:wmat} is the identity matrix and our proposed ADMM updates in Section~\ref{subsec:tvdn} in fact converge in one iteration.} so we instead refer to this proximal operator implicitly as the solution to the (weighted) total variation denosing (TVDN) problem
\begin{equation} \label{eq:tvdn}
\TVDN_{\lambda}(Z) := \arg \min_S ~ \tfrac{1}{2} \|Z - S\|_F^2 + \lambda\WTV(S).
\end{equation}
Using the above results and notation, we can express the proximal updates \eqref{eq:prox:updates:1} as
\begin{equation} \label{eq:prox:updates:2}
\begin{aligned}
U^{k+1} &= \proj_{M}(L^{k} + S^{k} + E^{k} - Y) \\[4pt]
L^{k+1} &= \SVT_{\tau^k \lambda_L}(L^k - \tau^k U^{k+1}) \\[4pt]
S^{k+1} &= \TVDN_{\tau^k \lambda_S}(S^k - \tau^k U^{k+1}) \\[4pt]
E^{k+1} &= \soft_{\tau^k \lambda_E}(E^k - \tau^k U^{k+1}),
\end{aligned}
\end{equation}
where it remains to describe how to compute $S^{k+1}$.

\subsection{Total Variation Denoising Updates} \label{subsec:tvdn}
Using the notation from Section~\ref{subsec:tv}, we can equivalently express the operator $\TVDN_{\lambda}(Z)$ as the solution to the vector-valued problem
\begin{equation} \label{eq:tvdn:2}
\min_{s} ~ \tfrac{1}{2}\|z - s\|_2^2 + \lambda \|WCs\|_1,
\end{equation}
where $z = \vec(Z)$ and the matrices $W$ and $C$ are defined as in \eqref{eq:tv:vec}. We solve \eqref{eq:tvdn:2} using the alternating direction method of multipliers (ADMM) \cite{boyd2011distributed}, a powerful general-purpose method for minimizing convex problems of the form $f(x) + g(x)$ subject to linear equality constraints. To apply ADMM, we perform the variable split $v = Cs$ and write \eqref{eq:tvdn:2} as the equivalent constrained problem
\begin{equation} \label{eq:tvdn:3}
\begin{array}{r@{~}l}
\displaystyle\min_{s,v} & ~ \tfrac{1}{2}\|z - s\|_2^2 + \lambda \|Wv\|_1 \\[6pt]
\text{s.t.} & ~ Cs - v = 0,
\end{array}
\end{equation}
which is in the standard form for ADMM.\footnote{Note that we choose the split $v = Cs$ rather than the split $v = WCs$ because the resulting ADMM updates in the former case have efficient closed-form solutions that leverage the block-circulant structure of $C$ \eqref{eq:cmat}.} The ADMM updates for \eqref{eq:tvdn:3} are
\begin{equation} \label{eq:tvdn:admm}
\begin{aligned}
s^{k+1} =& ~ \arg \min_{s} ~ \tfrac{1}{2} \|z-s\|_2^2 + \tfrac{\rho}{2}\|Cs-v^k+u^k\|_2^2 \\[2pt]
v^{k+1} =& ~ \arg \min_{v} ~ \lambda\|Wv\|_1 + \tfrac{\rho}{2}\|Cs^{k+1}-v+u^k\|_2^2 \\[2pt]
u^{k+1} =& ~ u^k+Cs^{k+1}-v^{k+1}
\end{aligned}
\end{equation}
with parameter $\rho > 0$. The $s$ update in \eqref{eq:tvdn:admm} is a least squares problem with normal equation
\begin{equation} \label{eq:tvdn:s:normeq}
(I + \rho C^TC)s^{k+1} = z + \rho C^T(v^k - u^k),
\end{equation}
so the solution could in principal be obtained by computing the matrix inverse $(I + \rho C^TC)^{-1}$. However, this matrix has a special block-circulant structure that admits a fast closed-form solution using fast Fourier transforms (FFTs). Indeed, the exact solution can be computed \cite{cao2016total,tao2009alternating} as
\begin{equation} \label{eq:tvdn:s:fft}
s^{k+1} = \mathcal{F}_3^{-1}\left[\frac{\mathcal{F}_3(z + \rho C^T(v^k-u^k))}{1 + \rho  \mathcal{F}_3(c)}\right],
\end{equation}
where $\mathcal{F}_3 : \R^{mnp} \rightarrow \R^{mnp}$ denotes the operator that reshapes its input into an $m \times n \times p$ tensor, computes the 3D Fourier transform, and vectorizes the result; $c$ is the first column of $C^TC$; and division is performed elementwise. The denominator of \eqref{eq:tvdn:s:fft} is a constant and can be precomputed.

The vector $c \in \R^{mnp}$ has special structure. Indeed, one can show that
\begin{equation}
c = \vec(|\mathcal{F}_1(d_m)|^2 \circ |\mathcal{F}_1(d_n)|^2 \circ |\mathcal{F}_1(d_p)|^2),
\end{equation}
where $\mathcal{F}_1(\cdot)$ denotes the 1D Fourier transform of a vector; $\left|\cdot\right|^2$ denotes elementwise squared-magnitude; the vector $d_n = [-1~0~\ldots~0~1]^T \in \R^n$ is the first column of \eqref{eq:Dn}; and $T = a \circ b \circ c$ is the order three tensor sum of vectors $a$, $b$, and $c$---\ie the tensor with entries $T_{ijk} = a_i + b_j + c_k$.

The $W$ matrix in the $v$-update of \eqref{eq:tvdn:admm} is a diagonal matrix, so the $v$ update has a simple closed-form solution involving elementwise soft-thresholding with an entry-dependent threshold, which we write as
\begin{equation} \label{eq:tvdn:v}
v^{k+1} = \soft(Cx^{k+1}+u^k,~(\lambda/\rho)w),
\end{equation}
where $\soft(x,y) = \sign(x) \odot (x - y)_+$ is interpreted elementwise for vectors and $w$ is the main diagonal of $W$.

\subsection{Improved Low-Rank Update} \label{subsec:optshrink}
Motivated by recent work \cite{moore2014improved,ravishankar2017lassi}, we propose to replace the SVT operator in the $L$ update of \eqref{eq:prox:updates:2} with an improved low-rank matrix estimator (OptShrink) \cite{nadakuditi2014optshrink} that has been shown to produce superior low-rank components in practice. Our proposed (modified) update scheme thus becomes
\begin{equation} \label{eq:prox:update:optshrink}
\begin{array}{r@{~}c@{~}l}
U^{k+1} &=& \proj_{M}(L^{k} + S^{k} + E^{k} - Y) \\[4pt]
L^{k+1} &=& \OptShrink_r\left(L^k - \tau^k U^{k+1}\right) \\[4pt]
S^{k+1} &=& \TVDN_{\tau^k \lambda_S}(S^k - \tau^k U^{k+1}) \\[4pt]
E^{k+1} &=& \soft_{\tau^k \lambda_E}(E^k - \tau^k U^{k+1}).
\end{array}
\end{equation}
In \eqref{eq:prox:update:optshrink}, $\OptShrink(\cdot)$ is the low-rank matrix estimator defined for a given parameter $r > 0$ as
\begin{equation} \label{eq:optshrink}
\OptShrink_r(Z) = \sum_{i=1}^{r}\left(-2\frac{D_{\mu_Z}(\sigma_i)}{D_{\mu_Z}^\prime(\sigma_i)}\right)u_i v_i^H,
\end{equation}
where $Z = \sum_i \sigma_i u_i v_i^T$ is the SVD of $Z \in \R^{a \times b}$. In \eqref{eq:optshrink}, the $D$-transform is defined for a given probability measure $\mu$ as
\begin{equation} \label{eq:dtrans}
\begin{array}{r@{~}l}
D_{\mu}(z) = & \bigg[ \displaystyle\int \frac{z}{z^2 - t^2} \ \mathrm{d}\mu(t) \bigg] \times \\[10pt] & \qquad \bigg[ c \displaystyle\int \frac{z}{z^2 - t^2} \ \mathrm{d}\mu(t) + \frac{1 - c}{z} \bigg],
\end{array}
\end{equation}
where $D_{\mu}'(z)$ is the derivative of $D_{\mu}(z)$ with respect to $z$, $c = \min(a,b)/\max(a,b)$, and $\mu_{Z}(t) = \frac{1}{q-r} \sum_{i=r+1}^{q} \delta(t - \sigma_i)$
is the empirical mass function of the noise-only singular values of $Z$ with $q = \min(a,b)$. Note that the integrals in the $D$-transform terms in \eqref{eq:optshrink} reduce to summations for this choice of ${\mu}_{Z}$, so they can be computed efficiently.

The $\OptShrink_r(Z)$ operator computes the rank $r$ truncated SVD of $Z$ and then applies the shrinkage function defined by the parenthesized term in \eqref{eq:optshrink} to the leading singular values. We refer to the $D$-transform term as a shrinkage function because it shrinks its argument towards zero \cite{nadakuditi2014optshrink}. In contrast, the $\SVT_{\lambda}(Z)$ operator \eqref{eq:svt} applies a constant shrinkage level $\lambda$ to all singular values.

The OptShrink estimator provides two key benefits over SVT. First, it applies a data-driven shrinkage to the singular value spectrum of its argument, the form of which is imputed from the non-leading (noise) singular values. Generically, a smaller shrinkage is applied to larger---and hence more-informative---singular values and a comparatively larger shrinkage to smaller singular values. The effect of this nonlinear shrinkage is to produce an improved estimate of the underlying low-rank matrix embedded in the data \cite{nadakuditi2014optshrink}. See Appendix~\ref{app:optshrink:background} for further details. Second, OptShrink has a single parameter $r$ that directly specifies the rank of the output matrix. In the context of this work, it is very natural to set the rank parameter. Indeed, since our data $Y$ from \eqref{eq:Ymatrix} is registered, we can model the background of the registered video as static. In this case, the low-rank component $L$ of our model \eqref{eq:model} should ideally be a rank-$1$ matrix whose columns are repeated (up to scaling) vectorized copies of the static background image. In practice, the registered background may not be perfectly static, but it will still have high temporal correlation, so a small rank ($r = 2,3,\ldots$) will often suffice.

Note that the shrinkage applied by the OptShrink operator is data-driven, \ie it depends on singular values $r+1$ and above of its argument. As such, OptShrink does not correspond to the proximal operator of a penalty function $\phi(L)$,\footnote{In constrast, the $\SVT_{\lambda}(\cdot)$ operator applies soft thresholding shrinkage with parameter $\lambda$ to each singular value, which corresponds to the absolute value penalty $\phi(t) = \lambda|t|$.} so the updates \eqref{eq:prox:update:optshrink} are not proximal gradient updates for a cost function like \eqref{eq:cost}. Nonetheless, recent alternating minimization schemes involving OptShrink \cite{moore2014improved,ravishankar2017lassi} have proven to be numerically stable and yield convergent iterate sequences, and our numerical experiments in Section~\ref{sec:experiments} corroborate these findings.

Algorithm~\ref{alg:proposed} summarizes the proposed algorithm with OptShrink-based low-rank update. Henceforward, we refer to our method as Panoramic Robust PCA (PRPCA).

\begin{algorithm}[t!]
\begin{algorithmic}
\State \textbf{Input:} Video frames $F_1$, $\dots$, $F_p$ and parameters $r > 0$, $\lambda_S > 0$, $\lambda_E > 0$, $\tau < 2/3$, $\rho > 0$, and $K > 0$.
\State Compute registered frames $\widetilde{F}_1 \dots \widetilde{F}_p$ via \eqref{eq:registration}.
\State Construct $Y$ and $M$ matrices via \eqref{eq:Ymatrix}.
\State \textbf{Initialization:} $L^0 = U^0 = \proj_M(Y)$, $S^0 = E^0 = 0$, and $k = 0$.
\While {not converged}
\State $k = k + 1$
\State Update $U^k$, $L^k$, and $S^k$ via \eqref{eq:prox:update:optshrink}
\State Update $S^k$ by performing $K$ iterations of \eqref{eq:tvdn:admm}
\EndWhile
\State \textbf{Output:} Decomposition $\{L^k ,\, S^k ,\, E^k\}$
\end{algorithmic}
\caption{Proposed PRPCA Algorithm}
\label{alg:proposed}
\end{algorithm}

\subsection{Complexity Analysis} \label{subsec:complexity}
We now analyze the computational complexity of our PRPCA method from Algorithm~\ref{alg:proposed}. For each outer iteration, the $U$ and $E$ updates require $O(mnp)$ operations, and the cost of computing the $L$ update is $O(m^2n^2p)$---the cost of computing the SVD of a tall $mn \times p$ matrix \cite{golub1970singular}. Finally, the cost of updating $E$ using the ADMM-based scheme \eqref{eq:tvdn:admm} is $O(Kmnp\log(mnp))$, where $K$ is the number of ADMM iterations applied and the per-iteration cost is determined by the cost of computing a 3D FFT of an $m \times n \times p$ tensor \cite{cooley1965algorithm}. Therefore the overall per-iteration cost of our proposed algorithm is dominated by the cost of computing the SVD of a $mn \times p$ matrix, which is the same complexity as RPCA, TVRPCA, and most other robust PCA algorithms involving rank penalties.

In practice, moving camera video magnifies the size of the registered data $Y$ processed by our algorithm compared to the data matrices of the other methods. Since the complexity is quadratic in the number of pixels, a twofold increase in pixels (substantial camera motion) would make our algorithm roughly four-times slower than the other methods.

\section{Numerical Experiments} \label{sec:experiments}
Although our proposed PRPCA method is able to handle arbitrary camera motion due to the inclusion of the frame registration preprocessing step, the special case of robust foreground-background separation on static camera video is also of great practical interest. Indeed, a variety of algorithms have been proposed in the literature to address the static camera setting, including a number of algorithms based on robust PCA-type models. As such, an important contribution of this work is to demonstrate that our proposed OptShrink-based low-rank model and our formulation of total variation-based foreground regularization are able to produce state-of-the-art results on static camera videos. We perform our evaluation by comparing to the recent RPCA \cite{candes2011robust}, TVRPCA \cite{cao2016total}, and DECOLOR \cite{zhou2013moving} methods on corrupted static camera videos. In addition, we provide numerical comparisons with GRASTA \cite{he2011online}, Prac-ReProCS \cite{guo2014online}, Layering Denoising \cite{guo2016video}, Fast PCP \cite{rodriguez2013fast}, noncvxRPCA \cite{kang2015robust}, MAMR \cite{ye2015foreground}, and ROSL \cite{shu2014robust}.

We then demonstrate the ability of our method to process corrupted moving camera videos, a scenario that few methods in the literature can handle. The RPCA and TVRPCA models explicitly employ a static camera model, so they are not applicable in the moving camera setting. The DECOLOR, GRASTA, Prac-ReProCS algorithms can, in principle, adapt to moving camera (i.e., dynamic subspaces) video, but, as we demonstrate, they are either not suitable for processing corrupted videos or their subspace tracking models are unable to accurately track the quickly evolving subspaces that arise from moving camera videos in practice. We also compare to the RASL \cite{peng2012rasl} and IncPCP-PTI \cite{chau2017panning} methods.

All methods under comparison are foreground-background separation methods, so they have components corresponding to the $L$ (background) and $S$ (foreground) components of our model. To facilitate a direct comparison, we repeat the cost functions of RPCA, TVRPCA, and DECOLOR from their respective papers here and rename the optimization variables so that the corresponding background and foreground components of each method are denoted by $L$ and $S$, respectively. In each case, we also use the matrix $Y$ to denote the matrix whose columns contain the vectorized frames of the (possibly corrupted) video.\footnote{Note that the other methods do not employ our frame registration preprocessing step, so here $Y$ contains the vectorized raw video frames.}

The RPCA \cite{candes2011robust,otazo2015low} method minimizes the cost
\begin{equation} \label{eq:rpca:main}
\min_{L,S} ~ \tfrac{1}{2} \|Y - L - S\|_F^2 + \lambda_L \|L\|_{\star} + \lambda_S\|S\|_1,
\end{equation}
where $L$ is the low-rank background component and $S$ is the sparse foreground component. The TVRPCA method minimizes the cost from Equation~(7) of \cite{cao2016total}, which, in our notation, is
\begin{equation} \label{eq:tvrpca:main}
\begin{array}{rl}
\displaystyle\min_{L,G,E,S} & \|L\|_{\star} + \lambda_1\|G\|_1 + \lambda_2\|E\|_1 + \lambda_3 \TV(S) \\[8pt]
\text{s.t.} & Y = L + G, ~ ~ G = E + S.
\end{array}
\end{equation}
In \eqref{eq:tvrpca:main}, $L$ is the low-rank background component and $G$ is a residual matrix that is further decomposed into a smooth foreground component $S$ and a sparse error term $E$. Here, we use $\TV(\cdot)$ to denote the standard (unweighted) anisotropic total variation penalty. The DECOLOR method minimizes the cost from Equation~(20) of \cite{zhou2013moving}, which, in our notation, is
\begin{equation} \label{eq:decolor:main}
\min_{\tau,L,S} ~ \tfrac{1}{2}\|\proj_{S^{\perp}}(Y \circ \tau - L)\|_F^2 + \alpha\|L\|_{\star} + \beta\|S\|_1 + \gamma \TV(S).
\end{equation}
In \eqref{eq:decolor:main}, $L$ is the low-rank (registered) background, $S_{ij} \in \{0,1\}$ is the (registered) foreground mask, $S^{\perp}$ is the orthogonal complement of $S$, $\tau$ are the 2D parametric transforms that register the input frames $Y$, and $\TV(\cdot)$ is again the standard (unweighted) anisotropic total variation penalty. Note that the DECOLOR method directly estimates the support of the foreground. Thus, to display a foreground component for DECOLOR, we plot $(Y - L \circ \tau^{-1}) \odot S$, the difference between the raw video and the estimated background restricted to the support of the estimated foreground mask.

\subsection{Static camera video} \label{subsec:staticcam}
We work with the I2R dataset\footnote{See http://perception.i2r.a-star.edu.sg/bk\_model/bk\_index.html.} of static camera sequences. The sequences contain between 523 and 3584 frames, each with a subset of 20 frames that have labeled (ground truth) foreground masks. We run each method on a subset of several hundred (contiguous) frames from each sequence containing 10 labeled frames. To evaluate the robustness of each method, we consider two corruption models: Gaussian noise (dense) and salt and pepper outliers (sparse).

To evaluate the denoising capabilities of each method, we measure the peak signal-to-noise ratio of the foreground (f-PSNR) and background (b-PSNR) pixels, respectively, in decibels (dB), using the ground truth foreground masks to distinguish between foreground and background.\footnote{We include PSNRs obtained by the popular video denoising method BM3D \cite{danielyan2012bm3d} in Tables~\ref{tab:i2r:outliers:bonus} and \ref{tab:i2r:gaussian:bonus}. However, note that the other methods under test in this section should not be evaluated solely based on denoising performance relative to BM3D because they have the additional mandate to decompose the scene into background and foreground.} We also measure the ability of each method to isolate the true foreground by thresholding the foreground component and computing the F-measure of these estimated masks with respect to the labeled masks.\footnote{For DECOLOR, we use the foreground mask returned by the algorithm.}


Each method contains various parameters that must be tuned in practice. In particular, the methods are iterative and we fix the number of outer iterations for each method to 150. We tune the remaining regularization parameters for each method independently on each dataset (no train-test-validation splits) by performing an exhaustive grid search over a wide range of parameter values (including any values recommended by the authors of the existing methods) to achieve a reasonable combination of good foreground denoising, background denoising, and F-measure performance for each algorithm on each dataset.\footnote{For the Fast PCP, noncvxRPCA, MAMR, and ROSL methods, we ran the implementations available in the LRS Library \cite{sobral2015lrs} with the default parameters provided by the software.}

In particular, for our proposed PRPCA method we chose the 3D version of the weighted TV penalty from \eqref{eq:tv}, and we used $K = 10$ inner ADMM updates with step size $\tau = 1/3$ and $\rho = 1$. We swept the regularization parameters $\lambda_S$ and $\lambda_E$ over multiple decades of values around $1/\sqrt{mn}$; in particular, we tried $\lambda_S = \kappa / \sqrt{mn}$ and $\lambda_E = \gamma / \sqrt{mn}$ for $\kappa,\gamma \in [10^{-5},~10^1]$. The optimal regularization values varied slightly for each dataset, but we found that the choices $\kappa = 0.01$ and $\gamma = 0.001$ are reasonable values to try on most datasets in practice, assuming the input frames are normalized to $[0,~1]$. For the OptShrink rank parameter $r$ we tried values $r=1,2,\ldots,6$, and the universal choice $r=1$ generally performed the best.

\begin{table*}[t!]
\resizebox{\textwidth}{!}{
\begin{tabular}{|c|ccc|ccc|ccc|ccc|}
\hline
\multirow{2}{*}{Sequence} &
\multicolumn{3}{c|}{Proposed} &
\multicolumn{3}{c|}{RPCA} &
\multicolumn{3}{c|}{TVRPCA} &
\multicolumn{3}{c|}{DECOLOR} \\
\cline{2-13}
& f-PSNR & b-PSNR & F-measure & f-PSNR & b-PSNR & F-measure & f-PSNR & b-PSNR & F-measure & f-PSNR & b-PSNR & F-measure \\
\hline \hline
Hall          & \textbf{38.94} & \textbf{37.98} & \textbf{0.60} & 27.12 & 32.63 & 0.19 & 36.50 &         37.42  & \textbf{0.60} & 27.02 & 31.63 & 0.17 \\
Fountain      & \textbf{39.73} & \textbf{35.48} & \textbf{0.74} & 26.99 & 32.06 & 0.21 & 36.87 & \textbf{35.48} &         0.72  & 26.89 & 30.69 & 0.15 \\
Escalator     & \textbf{33.15} & \textbf{31.56} & \textbf{0.72} & 23.45 & 26.27 & 0.35 & 30.91 &         30.96  &         0.69  & 23.27 & 22.17 & 0.25 \\
Water Surface & \textbf{42.14} & \textbf{36.96} & \textbf{0.94} & 22.92 & 31.45 & 0.40 & 40.14 &         36.81  &         0.82  & 22.12 & 20.66 & 0.26 \\
Shopping Mall & \textbf{40.26} &         39.83  & \textbf{0.74} & 25.06 & 34.62 & 0.31 & 37.43 & \textbf{40.88} &         0.73  & 25.01 & 31.42 & 0.26 \\
\hline \hline
Average       & \textbf{38.84} & \textbf{36.36} & \textbf{0.75} & 25.11 & 31.41 & 0.29 & 36.37 &         36.31  &         0.71  & 24.86 & 27.31 & 0.22 \\
\hline
\end{tabular}}
\caption{Performance metrics for each method on sequences from the I2R dataset corrupted by 20\% outliers.}
\label{tab:i2r:outliers}
\vspace{-0.1cm}
\end{table*}

\begin{table*}[t!]
\resizebox{\textwidth}{!}{
\begin{tabular}{|c|ccc|ccc|ccc|ccc|}
\hline
\multirow{2}{*}{Sequence} &
\multicolumn{3}{c|}{Proposed} &
\multicolumn{3}{c|}{RPCA} &
\multicolumn{3}{c|}{TVRPCA} &
\multicolumn{3}{c|}{DECOLOR} \\
\cline{2-13}
& f-PSNR & b-PSNR & F-measure & f-PSNR & b-PSNR & F-measure & f-PSNR & b-PSNR & F-measure & f-PSNR & b-PSNR & F-measure \\
\hline \hline
Hall          & \textbf{36.66} & \textbf{32.72} &         0.58  & 31.80 & 30.14 & 0.30 & 34.64 & 21.83 & \textbf{0.59} & 31.65 & 25.14 &         0.56  \\
Fountain      & \textbf{38.14} & \textbf{30.05} & \textbf{0.74} & 34.57 & 29.35 & 0.35 & 36.45 & 24.22 &         0.70  & 36.51 & 25.54 &         0.71  \\
Escalator     & \textbf{32.83} & \textbf{26.60} & \textbf{0.72} & 29.87 & 25.07 & 0.49 & 31.15 & 22.35 &         0.68  & 25.67 & 23.54 & \textbf{0.72} \\
Water Surface & \textbf{38.46} & \textbf{31.08} & \textbf{0.94} & 30.19 & 28.71 & 0.57 & 33.83 & 23.88 &         0.81  & 29.35 & 20.88 &         0.84  \\
Shopping Mall & \textbf{37.31} & \textbf{35.29} & \textbf{0.71} & 32.34 & 31.54 & 0.34 & 35.13 & 24.31 & \textbf{0.71} & 32.39 & 30.93 & \textbf{0.71} \\
\hline \hline
Average       & \textbf{36.68} & \textbf{31.15} & \textbf{0.74} & 31.75 & 28.96 & 0.41 & 34.24 & 23.32 &         0.70  & 31.11 & 25.21 &         0.71  \\
\hline
\end{tabular}}
\caption{Performance metrics for each method on sequences from the I2R dataset corrupted by 30 dB Gaussian noise.}
\label{tab:i2r:gaussian}
\vspace{-0.1cm}
\end{table*}

\begin{table*}[t!]
\resizebox{\textwidth}{!}{
\begin{tabular}{|c|c|ccc|ccc|ccc|ccc|}
\hline
\multicolumn{2}{|c|}{\multirow{2}{*}{Corruption}} &
\multicolumn{3}{c|}{Proposed} &
\multicolumn{3}{c|}{RPCA} &
\multicolumn{3}{c|}{TVRPCA} &
\multicolumn{3}{c|}{DECOLOR} \\
\cline{3-14}
\multicolumn{1}{|c}{} & \multicolumn{1}{c|}{}
& f-PSNR & b-PSNR & F-measure & f-PSNR & b-PSNR & F-measure & f-PSNR & b-PSNR & F-measure & f-PSNR & b-PSNR & F-measure \\
\hline \hline
\multirow{6}{*}{p}
  & 10\% & \textbf{41.48} & \textbf{39.37} & \textbf{0.60} & 30.35 & 32.67 & 0.27 & 38.38 & 38.98 & \textbf{0.60} & 30.28 & 31.54 & 0.29 \\
  & 20\% & \textbf{38.94} & \textbf{37.98} & \textbf{0.60} & 27.12 & 32.63 & 0.19 & 36.50 & 37.42 & \textbf{0.60} & 27.02 & 31.63 & 0.17 \\
  & 30\% & \textbf{37.69} & \textbf{36.21} & \textbf{0.59} & 25.40 & 32.39 & 0.15 & 34.94 & 36.08 &         0.58  & 30.27 & 31.54 & 0.29 \\
  & 40\% & \textbf{36.49} & \textbf{34.73} & \textbf{0.58} & 24.26 & 32.03 & 0.13 & 32.51 & 24.13 &         0.57  & 24.13 & 18.50 & 0.07 \\
  & 50\% & \textbf{35.84} & \textbf{33.73} & \textbf{0.57} & 23.57 & 31.49 & 0.12 & 29.85 & 18.11 &         0.49  & 23.47 & 14.61 & 0.07 \\
  & 60\% & \textbf{34.93} & \textbf{32.38} & \textbf{0.56} & 22.87 & 31.36 & 0.10 & 27.98 & 14.65 &         0.35  & 22.79 & 14.13 & 0.07 \\
\hline \hline
\multirow{6}{*}{SNR}
 & 5 dB  & \textbf{31.78} & \textbf{26.15} & \textbf{0.52} & 20.85 & 18.55 & 0.07 & 25.20 & 11.29 &         0.08  & 27.98 & 14.30 &         0.07  \\
 & 10 dB & \textbf{32.78} & \textbf{27.87} & \textbf{0.54} & 23.04 & 23.31 & 0.08 & 26.85 & 13.33 &         0.14  & 28.54 & 14.30 &         0.07  \\
 & 20 dB & \textbf{34.73} & \textbf{30.73} & \textbf{0.56} & 27.42 & 28.73 & 0.14 & 30.20 & 16.89 &         0.34  & 30.13 & 14.30 &         0.07  \\
 & 30 dB & \textbf{36.66} & \textbf{32.72} &         0.58  & 31.80 & 30.14 & 0.30 & 34.64 & 21.83 & \textbf{0.59} & 31.65 & 25.14 &         0.56  \\
 & 40 dB & \textbf{39.64} & \textbf{33.90} & \textbf{0.60} & 36.20 & 31.27 & 0.46 & 37.96 & 25.70 &         0.58  & 36.27 & 31.51 &         0.59  \\
 & 50 dB & \textbf{42.89} & \textbf{36.14} &         0.60  & 40.59 & 32.00 & 0.54 & 41.47 & 29.77 &         0.59  & 37.87 & 32.73 & \textbf{0.61} \\
\hline
\end{tabular}}
\caption{Performance metrics on the Hall sequence as a function of outlier probability $p$ and SNR (Gaussian noise).}
\label{tab:hall:outliers:gaussian}
\vspace{-0.1cm}
\end{table*}

\begin{table*}[t!]
\begin{center}
\resizebox{1.00\textwidth}{!}{
\begin{tabular}{|c|ccc|ccc|ccc|ccc|}
\hline
\multirow{2}{*}{Sequence} &
\multicolumn{3}{c|}{Layering Denoising} &
\multicolumn{3}{c|}{GRASTA} &
\multicolumn{3}{c|}{Prac-ReProCS} &
\multicolumn{3}{c|}{BM3D} \\
\cline{2-13}
& f-PSNR & b-PSNR & F-measure & f-PSNR & b-PSNR & F-measure & f-PSNR & b-PSNR & F-measure & f-PSNR & b-PSNR & F-measure \\
\hline \hline
Hall          &  32.92 & 23.16 & 0.06  &  26.72 & 25.18 & 0.15  &  21.30 & 30.80 & 0.14  &  34.96 & 25.77 & - \\
Fountain      &  33.46 & 23.61 & 0.07  &  26.78 & 27.92 & 0.17  &  23.09 & 29.55 & 0.14  &  35.71 & 24.72 & - \\
Escalator     &  27.02 & 20.58 & 0.12  &  23.50 & 13.37 & 0.20  &  18.78 & 24.88 & 0.33  &  28.53 & 21.93 & - \\
Water Surface &  32.64 & 26.72 & 0.15  &  23.01 & 23.98 & 0.23  &  22.40 & 31.51 & 0.38  &  34.43 & 27.84 & - \\
Shopping Mall &  32.52 & 25.98 & 0.10  &  24.98 & 19.61 & 0.17  &  21.32 & 32.96 & 0.20  &  34.19 & 28.25 & - \\
\hline \hline
Average       &  31.71 & 24.01 & 0.10  &  25.00 & 22.01 & 0.18  &  21.38 & 29.94 & 0.24  &  33.56 & 25.70 & - \\
\hline \hline
\multirow{2}{*}{Sequence} &
\multicolumn{3}{c|}{Fast PCP} &
\multicolumn{3}{c|}{noncvxRPCA} &
\multicolumn{3}{c|}{MAMR} &
\multicolumn{3}{c|}{ROSL} \\
\cline{2-13}
& f-PSNR & b-PSNR & F-measure & f-PSNR & b-PSNR & F-measure & f-PSNR & b-PSNR & F-measure & f-PSNR & b-PSNR & F-measure \\
\hline \hline
Hall          &  27.56 & 26.37 & 0.49  &  27.59 & 26.92 & 0.50  &  27.21 & 30.64 & 0.53  &  27.23 & 30.75 & 0.53 \\
Fountain      &  27.37 & 26.17 & 0.49  &  27.37 & 26.28 & 0.49  &  27.06 & 30.60 & 0.52  &  26.95 & 30.76 & 0.53 \\
Escalator     &  27.30 & 27.16 & 0.49  &  27.32 & 27.16 & 0.49  &  27.16 & 30.66 & 0.52  &  27.02 & 30.78 & 0.53 \\
Water Surface &  27.14 & 27.91 & 0.49  &  27.32 & 25.45 & 0.49  &  27.67 & 30.60 & 0.53  &  27.10 & 30.77 & 0.54 \\
Shopping Mall &  27.28 & 25.93 & 0.48  &  27.56 & 26.19 & 0.50  &  27.12 & 30.56 & 0.52  &  27.01 & 30.78 & 0.53 \\
\hline \hline
Average       &  27.33 & 26.71 & 0.49  &  27.43 & 26.40 & 0.49  &  27.24 & 30.61 & 0.52  &  27.06 & 30.77 & 0.53 \\
\hline
\end{tabular}}
\caption{Performance metrics for additional methods on the I2R dataset corrupted by 20\% outliers.}
\label{tab:i2r:outliers:bonus}
\end{center}
\vspace{-0.3cm}
\end{table*}

\begin{table*}[t!]
\begin{center}
\resizebox{1.00\textwidth}{!}{
\begin{tabular}{|c|ccc|ccc|ccc|ccc|}
\hline
\multirow{2}{*}{Sequence} &
\multicolumn{3}{c|}{Layering Denoising} &
\multicolumn{3}{c|}{GRASTA} &
\multicolumn{3}{c|}{Prac-ReProCS} &
\multicolumn{3}{c|}{BM3D} \\
\cline{2-13}
& f-PSNR & b-PSNR & F-measure & f-PSNR & b-PSNR & F-measure & f-PSNR & b-PSNR & F-measure & f-PSNR & b-PSNR & F-measure \\
\hline \hline
Hall          &  36.64 & 28.45 & 0.06  &  32.02 & 27.92 & 0.13  &  21.19 & 22.97 & 0.12  &  38.82 & 31.99 & - \\
Fountain      &  37.99 & 29.97 & 0.06  &  34.58 & 26.31 & 0.15  &  24.33 & 24.20 & 0.14  &  40.52 & 31.13 & - \\
Escalator     &  31.10 & 25.41 & 0.12  &  31.03 & 14.14 & 0.22  &  19.26 & 20.57 & 0.28  &  32.49 & 28.52 & - \\
Water Surface &  37.48 & 30.28 & 0.51  &  30.46 & 26.78 & 0.22  &  23.43 & 23.05 & 0.33  &  41.36 & 32.18 & - \\
Shopping Mall &  38.69 & 32.15 & 0.10  &  32.61 & 27.99 & 0.21  &  22.07 & 24.91 & 0.19  &  39.78 & 34.89 & - \\
\hline \hline
Average       &  36.38 & 29.25 & 0.17  &  32.14 & 24.63 & 0.19  &  22.06 & 23.14 & 0.21  &  38.59 & 31.74 & - \\
\hline \hline
\multirow{2}{*}{Sequence} &
\multicolumn{3}{c|}{Fast PCP} &
\multicolumn{3}{c|}{noncvxRPCA} &
\multicolumn{3}{c|}{MAMR} &
\multicolumn{3}{c|}{ROSL} \\
\cline{2-13}
& f-PSNR & b-PSNR & F-measure & f-PSNR & b-PSNR & F-measure & f-PSNR & b-PSNR & F-measure & f-PSNR & b-PSNR & F-measure \\
\hline \hline
Hall          &  27.83 & 29.94 & 0.51  &  27.63 & 29.88 & 0.50  &  27.81 & 29.38 & 0.50  &  27.45 & 30.19 & 0.51 \\
Fountain      &  27.83 & 29.85 & 0.51  &  28.00 & 29.71 & 0.48  &  27.74 & 29.39 & 0.51  &  27.78 & 30.21 & 0.50 \\
Escalator     &  27.79 & 29.74 & 0.51  &  27.86 & 29.72 & 0.49  &  28.08 & 29.41 & 0.50  &  27.81 & 30.16 & 0.52 \\
Water Surface &  27.97 & 29.87 & 0.50  &  27.82 & 29.75 & 0.49  &  27.79 & 29.38 & 0.50  &  27.62 & 30.13 & 0.50 \\
Shopping Mall &  27.87 & 29.88 & 0.49  &  28.05 & 29.77 & 0.48  &  27.66 & 29.38 & 0.52  &  27.73 & 30.21 & 0.50 \\
\hline \hline
Average       &  27.86 & 29.86 & 0.50  &  27.87 & 29.77 & 0.49  &  27.82 & 29.39 & 0.51  &  27.68 & 30.18 & 0.51 \\
\hline
\end{tabular}}
\caption{Performance metrics for additional methods on the I2R dataset corrupted by 30 dB Gaussian noise.}
\label{tab:i2r:gaussian:bonus}
\end{center}
\vspace{-0.3cm}
\end{table*}

\begin{figure*}[t!]
\centering\includegraphics[width=0.95\textwidth]{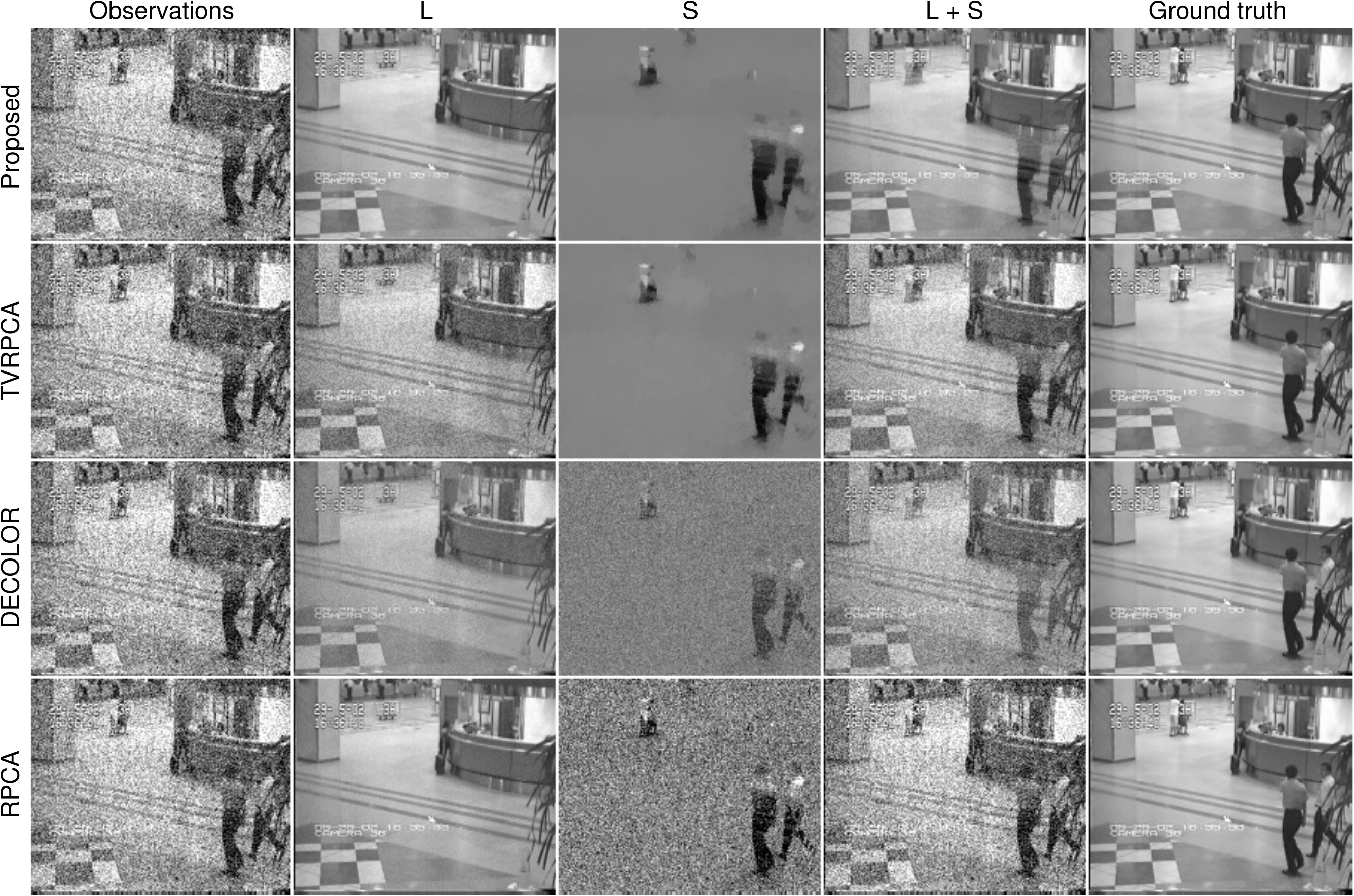}
\caption{A representative frame from the decompositions produced by each method applied to the Hall sequence corrupted by 30 dB Gaussian noise. Left column: observations; $L$: reconstructed background; $S$: reconstructed foreground; $L+S$: reconstructed scene; right column: Hall sequence.}
\label{fig:hall:gaussian}
\end{figure*}

\begin{figure*}[t!]
\centering\includegraphics[width=\textwidth]{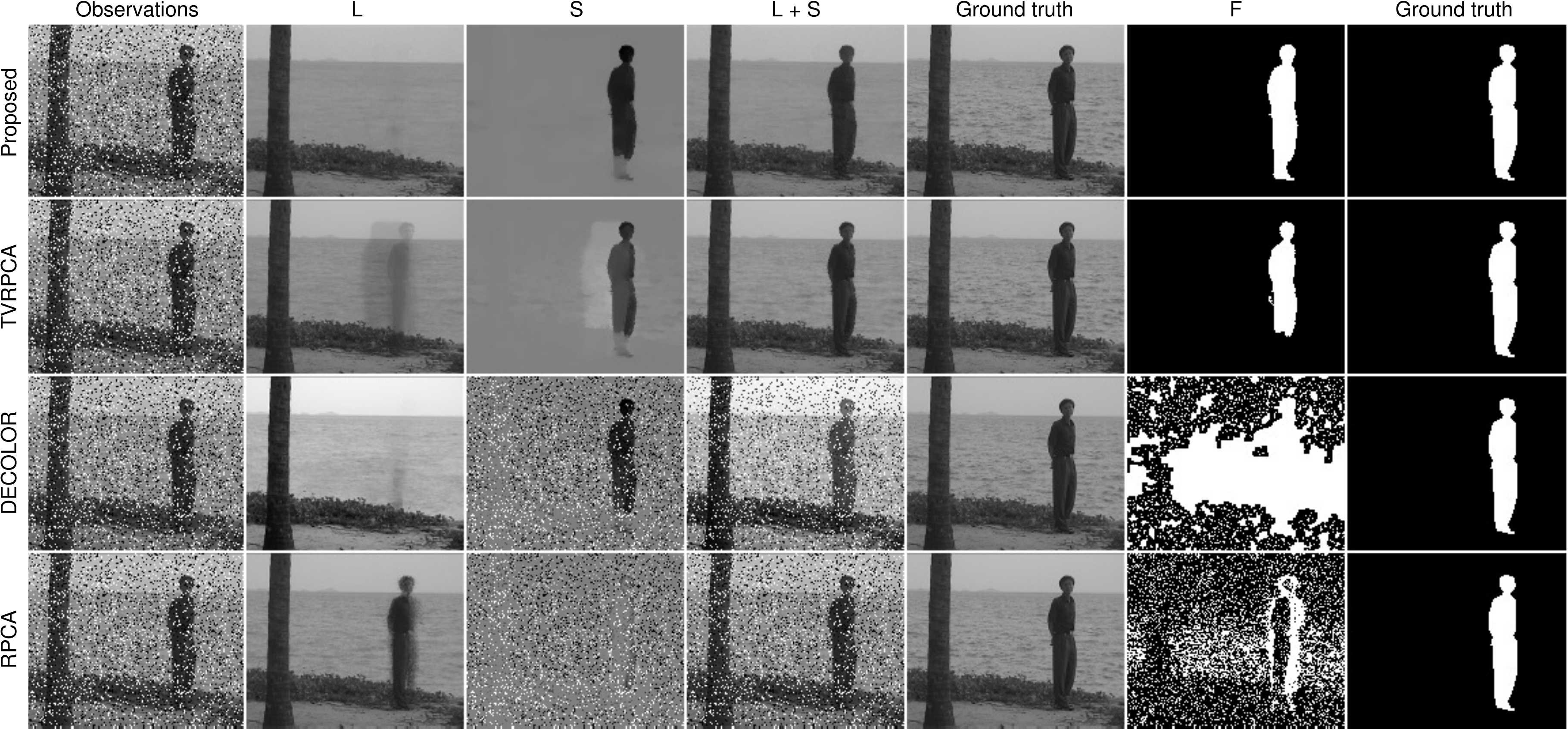}
\caption{A representative frame from the decompositions produced by each method applied to the Water Surface sequence corrupted by 20\% outliers. Left column: observations; $L$: reconstructed background; $S$: reconstructed foreground; $L+S$: reconstructed scene; fifth column: Water Surface sequence; $F$: estimated foreground mask; right column: true mask.}
\label{fig:water:outliers}
\end{figure*}

Tables~\ref{tab:i2r:outliers} and \ref{tab:i2r:gaussian} compare the performance of the proposed method, TVRPCA, DECOLOR, and RPCA on the I2R sequences corrupted by 20\% salt and pepper outliers and Gaussian noise with 30 dB SNR, respectively. In addition, Table~\ref{tab:hall:outliers:gaussian} compares the performance of each method on the Hall sequence as a function of outlier probability and Gaussian noise SNR. Finally, Tables~\ref{tab:i2r:outliers:bonus} and \ref{tab:i2r:gaussian:bonus} provide additional comparisons with Layering Denoising, GRASTA, Prac-ReProCS, BM3D, Fast PCP, noncvxRPCA, MAMR, and ROSL on the experiments from Tables~\ref{tab:i2r:outliers} and \ref{tab:i2r:gaussian}.\footnote{Note that BM3D is a video denoising method, not a robust foreground-background separation method, so it has no associated F-measure.} Clearly, our proposed method performs significantly better than the existing methods across datasets, corruption type, and corruption level in nearly all cases.

Figures~\ref{fig:hall:gaussian} and \ref{fig:water:outliers} illustrate the decompositions produced by each method on the Hall sequence corrupted by 30 dB Gaussian noise and the Water Surface sequence corrupted by 20\% outliers. The foreground estimates of the RPCA and DECOLOR methods degrade dramatically when outliers are added because they lack the ability to distinguish outliers and other non-idealities from the underlying foreground component. TVPRCA performs better than these methods in the presence of outliers, but its estimated background component contains some residual dense corruptions (cf. Figure~\ref{fig:hall:gaussian}) and foreground artifacts (cf. Figure~\ref{fig:water:outliers}) that are not present in the proposed PRPCA method. Intuitively, the ghosting artifacts present in the background components produced by the TVRPCA and RPCA methods are due to the inability of the SVT-based updates to disentangle the dynamic foreground component of the scene from the background. See \cite{moore2014improved} for a more detailed investigation of the differences between SVT-based and OptShrink-based robust PCA algorithms. These results show that our proposed method is better able to uncover the true foreground and background components of corrupted video.

To compare the computational cost of each method, we measured the per-iteration runtime (averaged over 150 iterations) of each method on the Water Surface sequence. In our experiments, we used an Intel Core i5-5200U 2.20GHz processor. We found that the average per-iteration runtimes were 6.9 seconds for the proposed method, 4.2 seconds for TVRPCA, 2.0 seconds for DECOLOR, and 0.4 seconds for RPCA. The proposed method runs in comparable time to other state-of-the-art methods while producing substantially more accurate foreground and background estimates.

\begin{table*}[t!]
\centering\resizebox{\textwidth}{!}{
\begin{tabular}{|c|ccc|ccc|cc|c|c|c|c|}
\hline
\multirow{2}{*}{Sequence} &
\multicolumn{3}{c|}{Proposed} &
\multicolumn{3}{c|}{DECOLOR} &
\multicolumn{2}{c|}{Baseline (Median Filter)} &
Mask R-CNN & MIT & DeepLab & Deformable DL \\
\cline{2-13}
& f-PSNR & b-PSNR & F-measure & f-PSNR & b-PSNR & F-measure & f-PSNR & b-PSNR & F-measure & F-measure & F-measure & F-measure \\
\hline \hline
Tennis      & \textbf{39.39} & \textbf{30.13} & \textbf{0.76} &   -   &   -   &          -   & 38.58 & 27.36 & 0.75 & 0.38 & 0.07 & 0.00 \\
Paragliding & \textbf{42.26} & \textbf{33.86} & \textbf{0.78} & 26.07 & 18.93 &         0.44 & 41.54 & 29.17 & 0.29 & 0.13 & 0.00 & 0.00 \\
Rollerblade & \textbf{41.65} & \textbf{28.81} & \textbf{0.83} & 28.10 & 19.47 &         0.82 & 38.71 & 23.96 & 0.13 & 0.24 & 0.00 & 0.00 \\
Horsejump   & \textbf{36.16} & \textbf{26.54} &         0.76  & 22.86 & 17.64 & \textbf{0.80} & 34.51 & 23.19 & 0.09 & 0.43 & 0.00 & 0.00 \\
\hline \hline
Average     & \textbf{39.86} & \textbf{29.84} & \textbf{0.78} & 25.68 & 18.68 &         0.69 & 38.34 & 25.92 & 0.32 & 0.30 & 0.02 & 0.00 \\
\hline
\end{tabular}}
\caption{Performance metrics for each method on sequences from the DAVIS dataset corrupted by 30\% outliers.}
\label{tab:davis:outliers}
\end{table*}

\begin{table*}[t!]
\centering\resizebox{\textwidth}{!}{
\begin{tabular}{|c|ccc|ccc|cc|c|c|c|c|}
\hline
\multirow{2}{*}{Sequence} &
\multicolumn{3}{c|}{Proposed} &
\multicolumn{3}{c|}{DECOLOR} &
\multicolumn{2}{c|}{Baseline (Wiener Filter)} &
Mask R-CNN & MIT & DeepLab & Deformable DL \\
\cline{2-13}
& f-PSNR & b-PSNR & F-measure & f-PSNR & b-PSNR & F-measure & f-PSNR & b-PSNR & F-measure & F-measure & F-measure & F-measure \\
\hline \hline
Tennis      & \textbf{38.61} & \textbf{27.86} &         0.75  & 21.94 & 16.87 & 0.36 & 35.40 & 25.56 & \textbf{0.87} &         0.69  & 0.77 & 0.46 \\
Paragliding & \textbf{40.64} & \textbf{32.60} & \textbf{0.76} & 28.54 & 19.17 & 0.24 & 37.60 & 27.76 &         0.01  & \textbf{0.76} & 0.00 & 0.00 \\
Rollerblade & \textbf{38.10} & \textbf{28.65} & \textbf{0.82} & 28.18 & 17.63 & 0.77 & 37.86 & 23.60 &         0.56  &         0.23  & 0.00 & 0.00 \\
Horsejump   & \textbf{34.38} & \textbf{26.98} &         0.74  & 23.10 & 16.64 & 0.76 & 33.23 & 23.34 & \textbf{0.80} &         0.60  & 0.05 & 0.00 \\
\hline \hline
Average     & \textbf{37.93} & \textbf{29.02} & \textbf{0.77} & 25.44 & 17.58 & 0.53 & 36.02 & 25.07 &         0.56  &         0.57  & 0.21 & 0.12 \\
\hline
\end{tabular}}
\caption{Performance metrics for each method on sequences from the DAVIS dataset corrupted by 10 dB Poisson noise.}
\label{tab:davis:noise}
\end{table*}

\begin{table*}[t!]
\centering\resizebox{0.65\textwidth}{!}{
\begin{tabular}{|c|ccc|ccc|cc|}
\hline
\multirow{2}{*}{SNR} &
\multicolumn{3}{c|}{Proposed} &
\multicolumn{3}{c|}{DECOLOR} &
\multicolumn{2}{c|}{Baseline (Wiener Filter)} \\
\cline{2-9}
& f-PSNR & b-PSNR & F-measure & f-PSNR & b-PSNR & F-measure & f-PSNR & b-PSNR \\
\hline
5 dB  & \textbf{35.10} & \textbf{27.00} & \textbf{0.74} & 22.78 & 18.38 & 0.10 &         35.05  & 23.93 \\
10 dB & \textbf{38.61} & \textbf{27.86} & \textbf{0.75} & 21.94 & 16.87 & 0.36 &         38.32  & 27.11 \\
15 dB &         40.45  & \textbf{30.54} & \textbf{0.76} & 21.70 & 17.14 & 0.42 & \textbf{41.02} & 29.18 \\
20 dB & \textbf{41.88} & \textbf{31.14} & \textbf{0.77} & 21.69 & 16.96 & 0.39 &         41.78  & 30.05 \\
\hline
\end{tabular}}
\caption{Performance metrics for each method on the Tennis sequence as a function of SNR (Poisson noise).}
\label{tab:tennis:poisson}
\vspace{-0.05in}
\end{table*}

\subsection{Moving camera video} \label{subsec:movingcam}
We next demonstrate the performance of our proposed PRPCA method on several moving camera sequences from the recent DAVIS benchmark dataset~\cite{Perazzi2016}. Each sequence has associated labeled foreground masks.

The RPCA and TVRPCA methods are not suitable for moving camera video, so we only consider the DECOLOR method. As in the static camera case, we consider both salt and pepper outliers (sparse) and Poisson noise (dense).
\footnote{We chose Poisson noise for the moving camera setting rather than Gaussian noise as we used in the static camera case to demonstrate that our model performs well even when the data-fit term does not have a maximum likelihood-type interpretation with respect to the additive noise. The results in this section demonstrate that the Gaussian-motivated data-fit term performs well even when the noise is Poisson, which suggests that there is likely little benefit in introducing the extra computational burden necessitated by the Poisson-likelihood into \eqref{eq:cost}.} Although the video registration procedure in Section~\ref{sec:registration} can handle corrupted data, we use the homographies computed from the original videos to isolate the influence of our proposed model \eqref{eq:cost} on reconstruction quality. We evaluate performance using the same error metrics and parameter tuning strategies from Section~\ref{subsec:staticcam}. In particular, we found that similar parameter values for our proposed method perform well in practice, with the exception that we adopted 2D TV because the camera motion reduces the temporal continuity of the foreground. To provide an additional benchmark for denoising quality, we also consider the PSNRs produced by the following baseline per-frame denoising methods: median filtering (outlier corruptions) and Wiener filtering (Poisson noise corruptions). Note that these baseline methods are not foreground-background separation strategies, so they have no associated F-measures.

\begin{table*}[t!]
\resizebox{1.0\textwidth}{!}{
\begin{tabular}{|c|ccc|ccc|ccc|ccc|}
\hline
\multirow{2}{*}{Sequence} &
\multicolumn{3}{c|}{GRASTA} &
\multicolumn{3}{c|}{Prac-ReProCS} &
\multicolumn{3}{c|}{RASL} &
\multicolumn{3}{c|}{IncPCP-PTI} \\
\cline{2-13}
& f-PSNR & b-PSNR & F-measure & f-PSNR & b-PSNR & F-measure & f-PSNR & b-PSNR & F-measure & f-PSNR & b-PSNR & F-measure \\
\hline \hline
Tennis        &  24.44 & 11.38 & 0.10  &  19.46 & 21.88 & 0.12  &  23.37 & 17.48 & 0.09  &  25.39 & 10.14 & 0.03 \\
Paragliding   &  27.35 & 11.40 & 0.06  &  21.44 & 24.95 & 0.09  &  25.97 & 17.89 & 0.05  &  27.71 & 13.82 & 0.02 \\
Rollerblade   &  25.23 & 11.03 & 0.08  &  18.42 & 22.25 & 0.12  &  25.16 & 21.09 & 0.07  &  25.61 & 13.32 & 0.03 \\
Horsejump     &  23.02 & 11.10 & 0.13  &  15.77 & 20.14 & 0.15  &  22.02 & 18.05 & 0.12  &  23.11 & 12.34 & 0.09 \\
\hline \hline
Average       &  25.01 & 11.23 & 0.09  &  18.77 & 22.31 & 0.12  &  24.13 & 18.63 & 0.08  &  25.46 & 12.41 & 0.04 \\
\hline
\end{tabular}}
\caption{Performance metrics for four additional methods on the DAVIS sequences corrupted by 30\% outliers.}
\label{tab:davis:outliers:bonus}
\end{table*}

\begin{table*}[t!]
\resizebox{1.0\textwidth}{!}{
\begin{tabular}{|c|ccc|ccc|ccc|ccc|}
\hline
\multirow{2}{*}{Sequence} &
\multicolumn{3}{c|}{GRASTA} &
\multicolumn{3}{c|}{Prac-ReProCS} &
\multicolumn{3}{c|}{RASL} &
\multicolumn{3}{c|}{IncPCP-PTI} \\
\cline{2-13}
& f-PSNR & b-PSNR & F-measure & f-PSNR & b-PSNR & F-measure & f-PSNR & b-PSNR & F-measure & f-PSNR & b-PSNR & F-measure \\
\hline \hline
Tennis        &  31.00 & 17.13 & 0.12  &  21.33 & 19.17 & 0.14  &  23.42 & 17.88 & 0.09  &  - & - & - \\
Paragliding   &  33.76 & 17.74 & 0.06  &  22.70 & 21.21 & 0.08  &  26.55 & 19.82 & 0.06  &  - & - & - \\
Rollerblade   &  30.85 & 15.76 & 0.08  &  18.85 & 19.02 & 0.13  &  25.64 & 21.02 & 0.08  &  26.10 & 15.97 & 0.02 \\
Horsejump     &  28.41 & 16.50 & 0.13  &  16.74 & 18.49 & 0.17  &  22.85 & 18.69 & 0.13  &  23.46 & 14.57 & 0.09 \\
\hline \hline
Average       &  31.00 & 16.78 & 0.10  &  19.91 & 19.47 & 0.13  &  24.62 & 19.35 & 0.09  &  24.78 & 15.27 & 0.06 \\
\hline
\end{tabular}}
\caption{Performance metrics for four additional methods on the DAVIS sequences corrupted by 10 dB Poisson noise.}
\vspace{-0.1in}
\label{tab:davis:noise:bonus}
\end{table*}

Tables~\ref{tab:davis:outliers} and \ref{tab:davis:noise} compare the performance of each method on DAVIS sequences corrupted by 30\% salt and pepper outliers and Poisson noise with 10 dB SNR, respectively. The tables also include F-measure values computed from the foreground masks generated by several recent deep learning-based semantic segmentation models. In particular, we considered a Mask R-CNN model \cite{he2017mask} trained on the Microsoft COCO dataset, the MIT Scene Parsing benchmark model \cite{zhou2017scene} trained on the MIT ADE20K dataset, a DeepLab model \cite{chen2018deeplab} trained on the PASCAL VOC dataset, and a Deformable DeepLab model \cite{dai2017deformable} trained on the Cityscapes dataset.\footnote{See Appendix~\ref{app:semantic:segmentation} for more details about the semantic segmentation models under test in this experiment.}
In addition, Table~\ref{tab:tennis:poisson} compares the performance of each method on the Tennis sequence as a function of SNR with Poisson noise. Finally, Tables~\ref{tab:davis:outliers:bonus} and \ref{tab:davis:noise:bonus} provide additional comparisons with the performance of GRASTA, Prac-ReProCS, RASL, and IncPCP-PTI on the experiments from Tables~\ref{tab:davis:outliers} and \ref{tab:davis:noise}. Our proposed method achieves consistently higher f-PSNR, b-PSNR, and F-measure in all cases, which suggests it is well-suited for processing a variety of corruption types and levels.

\begin{figure*}[t!]
\centering\includegraphics[width=\textwidth]{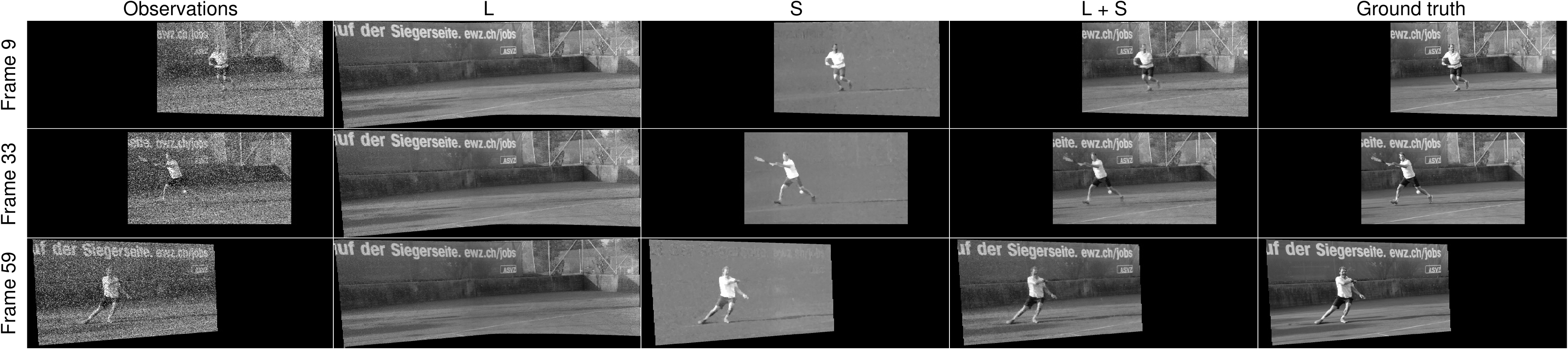}
\caption{Three representative frames produced by the proposed method on the Tennis sequence corrupted by 30\% salt and pepper outliers. Left column: registered observations; $L$: reconstructed registered background; $S$: reconstructed registered foreground; $L+S$: reconstructed registered scene restricted to the current field of view; right column: registered Tennis sequence.}
\label{fig:tennis:outliers:reg}
\vspace{-0.3cm}
\end{figure*}

\begin{figure*}[t!]
\centering\includegraphics[width=0.85\textwidth]{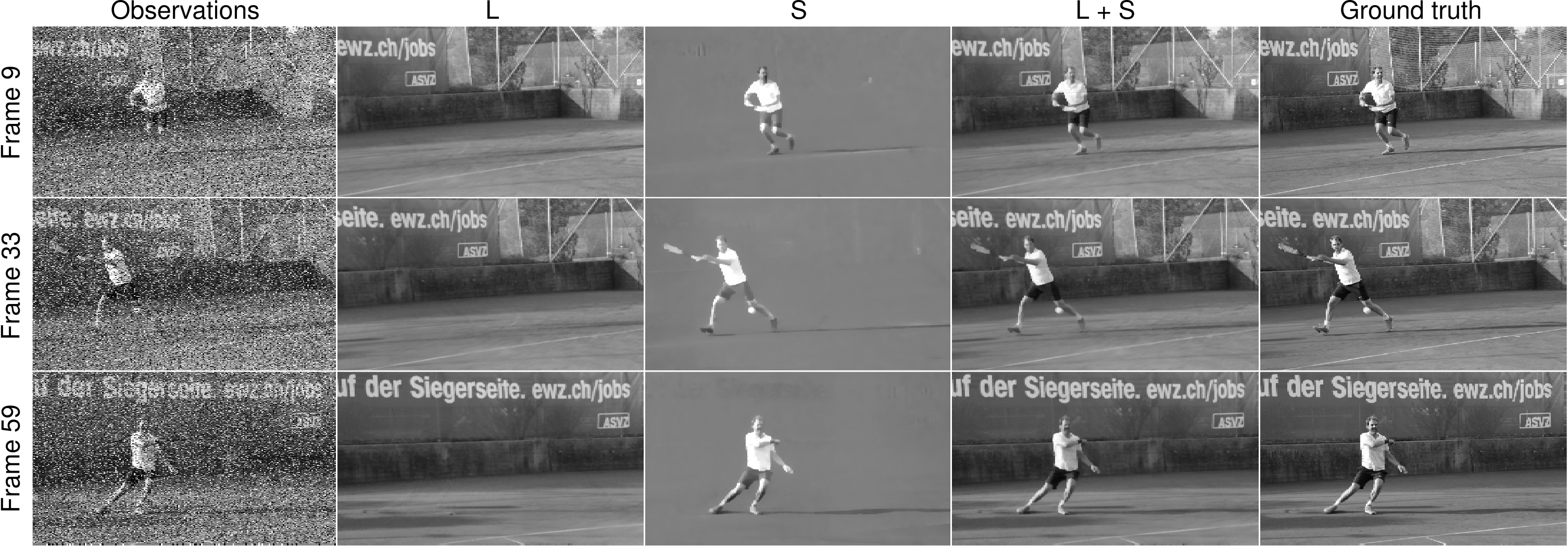}
\caption{The decompositions from Figure~\ref{fig:tennis:outliers:reg} mapped to the perspective of the original video.}
\label{fig:tennis:outliers}
\vspace{-0.1cm}
\end{figure*}

Figure~\ref{fig:tennis:outliers:reg} depicts the decompositions produced by our proposed PRPCA method on the Tennis sequence corrupted by 30\% salt and pepper outliers. Note how our proposed method gracefully aggregates the background information from the corrupted frames to produce a clean panoramic estimate ($L$) of the full field of view. Also, the registered TV-regularized component ($S$) is able to accurately estimate the dynamic foreground and decouple it from sparse corruptions. None of the methods considered in Section~\ref{subsec:staticcam} can produce comparable results. Figure~\ref{fig:tennis:outliers} shows the decompositions from Figure~\ref{fig:tennis:outliers:reg} mapped to the perspective of the original video by applying the inverse homographies computed during frame registration. These sequences constitute a direct decomposition of the original moving camera video.

\begin{figure*}[t!]
\begin{center}
\begin{subfigure}[b]{0.85\textwidth}
\centering\includegraphics[width=\textwidth]{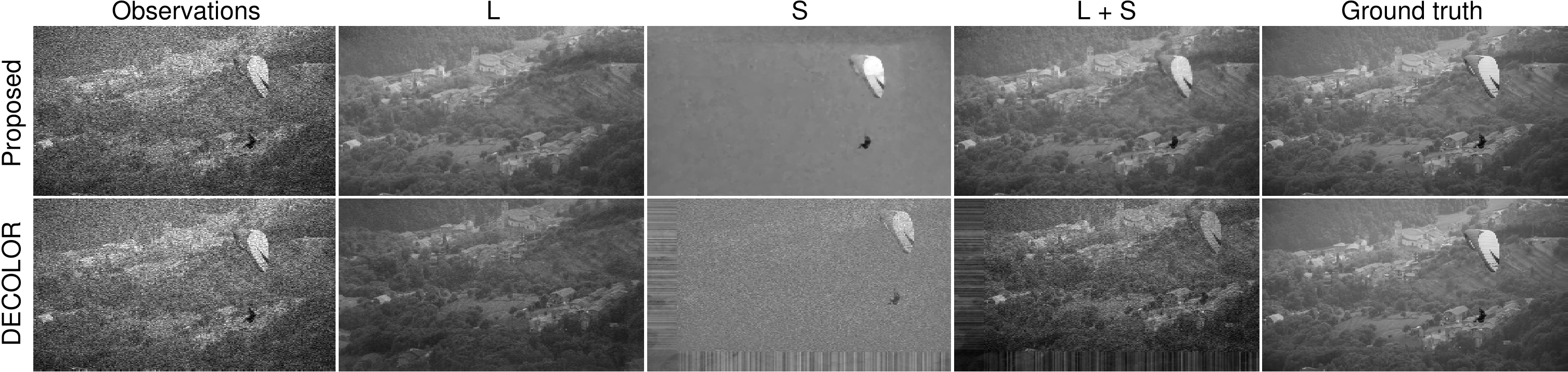}
\caption{Frame 10 of the Paragliding sequence.}
\vspace{0.15cm}
\end{subfigure}
\begin{subfigure}[b]{0.85\textwidth}
\centering\includegraphics[width=\textwidth]{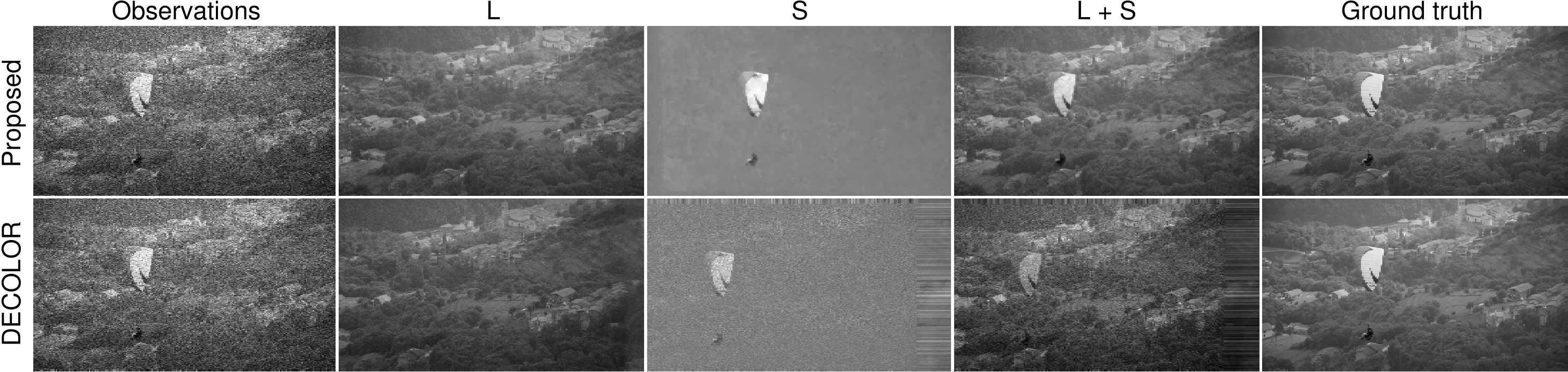}
\caption{Frame 34 of the Paragliding sequence.}
\end{subfigure}
\caption{Two representative frames from decompositions of the Paragliding sequence corrupted by 10 dB Poisson noise. Top row: decomposition produced by the proposed PRPCA method mapped to the perspective of the original video; bottom row: decomposition produced by DECOLOR. Left column: observations; $L$: reconstructed background; $S$: reconstructed foreground; $L+S$: reconstructed scene; right column: Paragliding sequence.}
\label{fig:paragliding:poisson}
\end{center}
\vspace{-0.2cm}
\end{figure*}

Figure~\ref{fig:paragliding:poisson} compares the performance of PRPCA and DECOLOR on the Paragliding sequence corrupted by 10 dB Poisson noise. DECOLOR fails to accurately estimate $L$ and $S$ due to the significant camera motion, while our proposed method consistently produces a high quality decomposition of the dynamic scene from the corrupted video.

\begin{figure}[t!]
\centering\includegraphics[width=\linewidth]{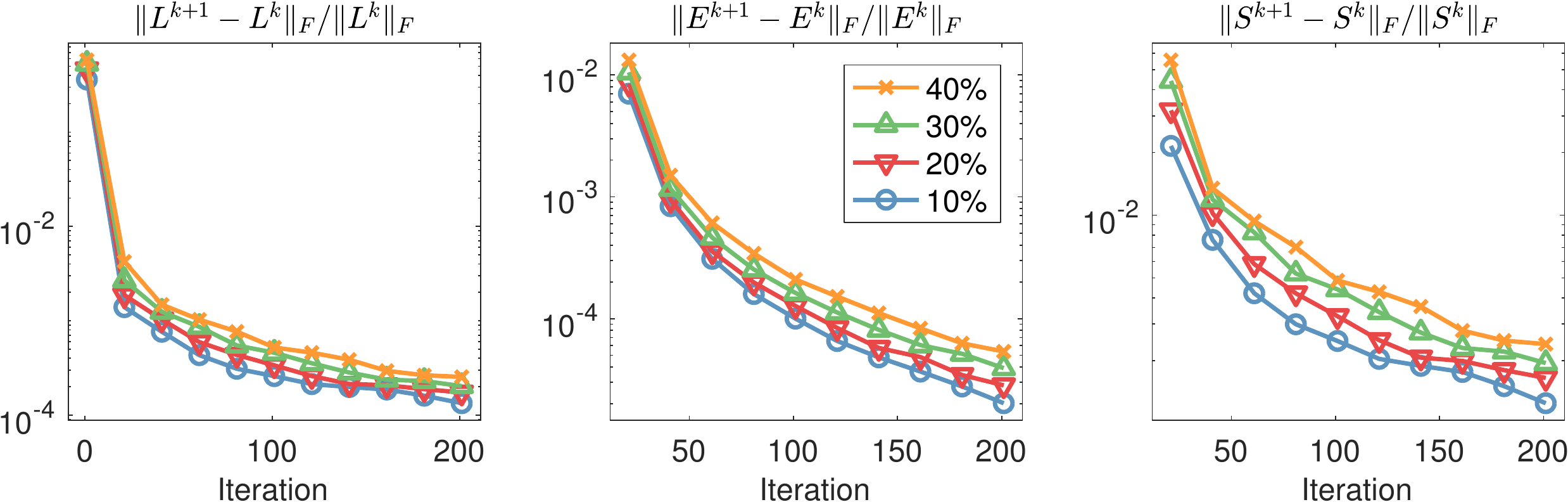}
\caption{Per-iteration convergence of the $L$, $E$, and $S$ components of the proposed PRPCA method on the Fountain sequence corrupted by outliers at various percentages.}
\label{fig:convergence}
\vspace{-0.2in}
\end{figure}

\begin{figure}[t!]
\centering\includegraphics[width=0.75\linewidth]{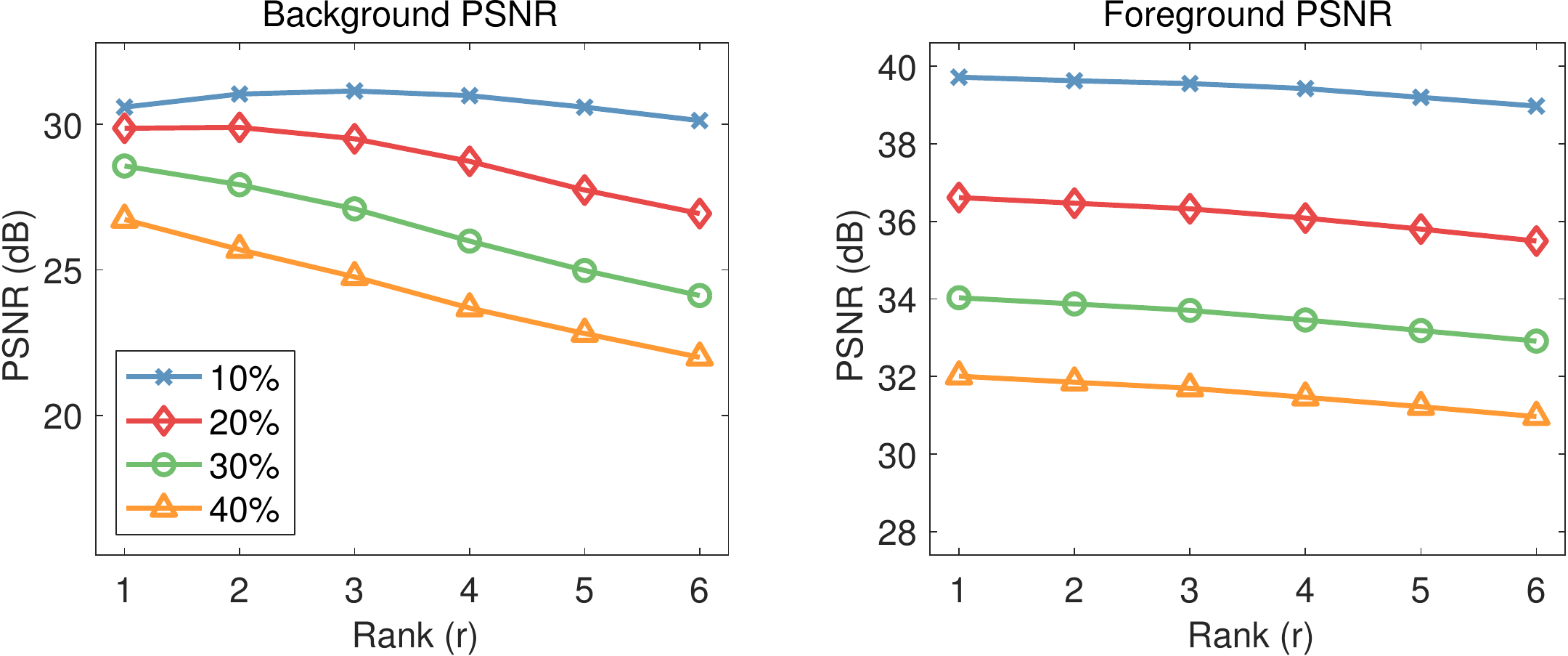}
\caption{Foreground and background PSNRs for the proposed method as a function of OptShrink rank parameter $r$ on the Tennis sequence for various percentages of outlier corruption.}
\label{fig:optshrink:psnr:rank}
\vspace{-0.2in}
\end{figure}

To compare the computational cost of each method on moving camera video, we measured the per-iteration runtime (averaged over 150 iterations) of each method on the Tennis sequence. In our experiments, we used an Intel Core i5-5200U 2.20GHz processor. For the proposed method, we found that the preprocessing (frame registration) step took 5.2 seconds, each outer iteration took an average of 65.4 seconds, and the post-processing (inverse registration) step took 8.8 seconds. For the DECOLOR method, we found that the preprocessing (registration) step took 71.7 seconds, each iteration took an average of 57.3 seconds, and the post-processing (mask warping) step took 2.6 seconds. These figures show that our proposed method runs in comparable time to the state-of-the-art DECLOR method while producing substantially more accurate foreground and background estimates from highly corrupted moving camera video.

\begin{figure}[t!]
\centering\includegraphics[width=0.75\linewidth]{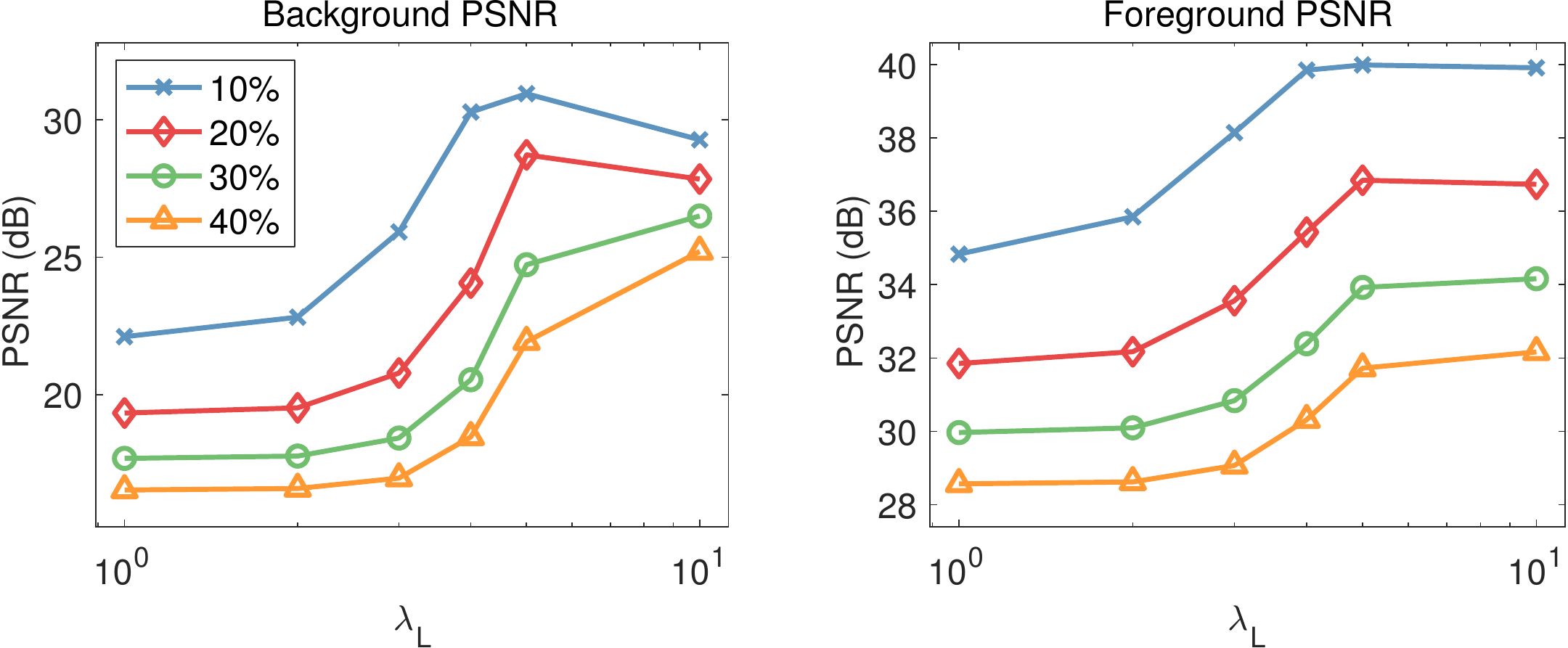}
\caption{Foreground and background PSNRs for the variation of the proposed method with SVT-based $L$ updates as a function of low-rank parameter $\lambda_L$ on the Tennis sequence for various percentages of outlier corruption.}
\label{fig:svt:psnr:lambdaL}
\end{figure}

\subsection{Algorithm properties} \label{subsec:properties}
In this section, we briefly investigate the properties of our PRPCA algorithm as described in Algorithm~\ref{alg:proposed}. Although the update scheme \eqref{eq:prox:update:optshrink} does not correspond to the proximal gradient updates of an explicit cost function that we can track, Figure~\ref{fig:convergence} demonstrates that the $L$, $S$, and $E$ iterates exhibit stable convergence behavior as the iterations progress.

It is interesting to understand how the performance of the OptShrink-based updates in Algorithm~\ref{alg:proposed} compares to the true proximal gradient approach from \eqref{eq:prox:updates:2} for minimizing \eqref{eq:cost}, which prescribes SVT-based $L$ updates. Figures~\ref{fig:optshrink:psnr:rank} and \ref{fig:svt:psnr:lambdaL} address this question by plotting the f-PSNRs and b-PSNRs produced by each method on the outlier-corrupted Tennis sequence as a function of their low-rank regularization parameters ($r$ in the case of our proposed method and $\lambda_L$ in the case of the SVT-based updates from \eqref{eq:prox:updates:2}). In each case, we set the foreground regularization parameters $\lambda_S$ and $\lambda_E$ to yield the best overall denoising performance (\ie to achieve the highest peaks in each figure, respectively). From Figure~\ref{fig:optshrink:psnr:rank} we see that our proposed method is well-behaved in the sense that the universal setting $r=1$ performs well regardless of corruption level; in addition, it is quite robust to rank overestimation in the sense that its performance degrades slowly as $r$ increases beyond its optimal value. Intuitively, this behavior is observed because the OptShrink estimator performs a data-driven shrinkage that minimizes the effect of superfluous rank components in $L$. On the other hand, the curves in Figure~\ref{fig:svt:psnr:lambdaL} are generally lower, which implies both that the optimal PSNR achieved by the SVT-based updates is usually lower and that the performance of the algorithm is more sensitive to the correct choice of $\lambda_L$, whose optimal value may vary in practice depending on the characteristics of the underlying data.

As previously mentioned, in our results thus far we use the homographies computed from the uncorrupted videos in our experiments to isolate the influence of our proposed model \eqref{eq:cost} on reconstruction quality. However, our video registration procedure from Section~\ref{sec:registration} is a robust RANSAC-based method that can also be applied to corrupted data. Indeed, Table~\ref{tab:tennis:homographies} demonstrates the performance of our proposed method as a function of outlier probability on the Tennis sequence when the homographies are computed from (a simple median-filtered version of) the corrupted video. The results indicate that our proposed method is not sensitive to small errors due to imperfect frame registration.

\begin{table}[t!]
\begin{center}
\resizebox{0.7\linewidth}{!}{
\begin{tabular}{|c|c|c|c|}
\hline
  p  & f-PSNR & b-PSNR & F-measure \\
\hline \hline
10\% & 42.12 & 30.28 & 0.76 \\
20\% & 41.50 & 30.47 & 0.75 \\
30\% & 39.06 & 29.19 & 0.75 \\
40\% & 37.62 & 28.44 & 0.73 \\
\hline
\end{tabular}}
\caption{Performance metrics for the proposed method on the Tennis sequence as a function of outlier corruption. For this experiment, the homographies for the frame registration procedure were computed from the corrupted video.}
\label{tab:tennis:homographies}
\end{center}
\vspace{-0.2in}
\end{table}

An interesting future line of inquiry is to investigate an extension of our proposed method where the frame registration procedure is included as an extra alternating step in our update scheme; such a variation would allow the homographies to be iteratively refined as the iterations progress (at the expense of additional computation), which would likely lead to more accurate frame registration and, hence, improved foreground and background estimates.

\section{Conclusion} \label{sec:conclusion}
We proposed a new panoramic robust PCA method for performing robust foreground-background separation on possibly corrupted video with arbitrary camera motion. Our proposed method registers the frames of the raw video, and it utilizes weighted total variation regularization and an improved low-rank matrix estimator (OptShrink) to jointly estimate the foreground and background components of the scene from the registered frames. Our numerical experiments demonstrate that our proposed method is robust to both dense and sparse corruptions of the raw video and produces superior foreground-background separations compared to existing methods. In future work, we plan to investigate the usefulness of the foreground components produced by our method for computer vision tasks like object tracking and activity detection.

\appendices

\section{Semantic Segmentation Models} \label{app:semantic:segmentation}
Here we provide more information about the four semantic segmentation models that we included in Tables~\ref{tab:davis:outliers} and \ref{tab:davis:noise}.

\subsection{Mask R-CNN}
The Mask R-CNN \cite{he2017mask} code that we used was downloaded from \url{https://github.com/roytseng-tw/Detectron.pytorch}. The codebase provides multiple model architectures, and we considered the model with \verb$ResNet-101-FPN$ backbone that was pre-trained on the Microsoft COCO dataset \cite{lin2014microsoft}.

\subsection{MIT Scene Parsing Benchmark}
The MIT Scene Parsing benchmark model \cite{zhou2017scene} code that we used was downloaded from \url{https://github.com/CSAILVision/semantic-segmentation-pytorch}. The codebase provides multiple model architectures, and we considered the model with \verb$ResNet50dilated$ encoder and \verb$PPM_deepsup$ decoder that was pre-trained on the MIT ADE20K dataset \cite{zhou2016semantic,zhou2017scene}.

\subsection{DeepLab}
The DeepLab \cite{chen2018deeplab} code that we used was downloaded from \url{https://github.com/DrSleep/tensorflow-deeplab-resnet/tree/tf-0.11}. The model was built on a \verb$ResNet-101$ backbone with dilated convolutions and was pre-trained on the PASCAL VOC dataset \cite{everingham2010pascal}.

\subsection{Deformable DeepLab}
The Deformable DeepLab \cite{dai2017deformable} code that we used was downloaded from \url{https://github.com/msracver/Deformable-ConvNets}. The codebase provides multiple model architectures, and we considered the model with \verb$ResNet-v1-101$ backbone that was trained on the Cityscapes dataset \cite{cordts2016cityscapes}.

\section{Additional Numerical Experiments} \label{app:additional:experiments}
Here we provide some additional numerical experiments that extend our investigation from Section~\ref{sec:experiments}.

Table~\ref{tab:i2r:missing} compares the performance of the proposed method, TVRPCA, DECOLOR, and RPCA on the I2R sequences corrupted by 70\% missing data. Note that, in the missing data case, it is trivial to incorporate the missing data locations in our model: we simply encode them as zeros in the mask matrix $M$. The RPCA, TVRPCA, and DECOLOR objectives do not directly support inpainting, but they can be easily modified to do so. See Appendix~\ref{app:missingdata} for a description of the modified versions of RPCA, TVRPCA, and DECOLOR that we used in our missing data experiments.

\begin{figure*}[t!]
\centering\includegraphics[width=\textwidth]{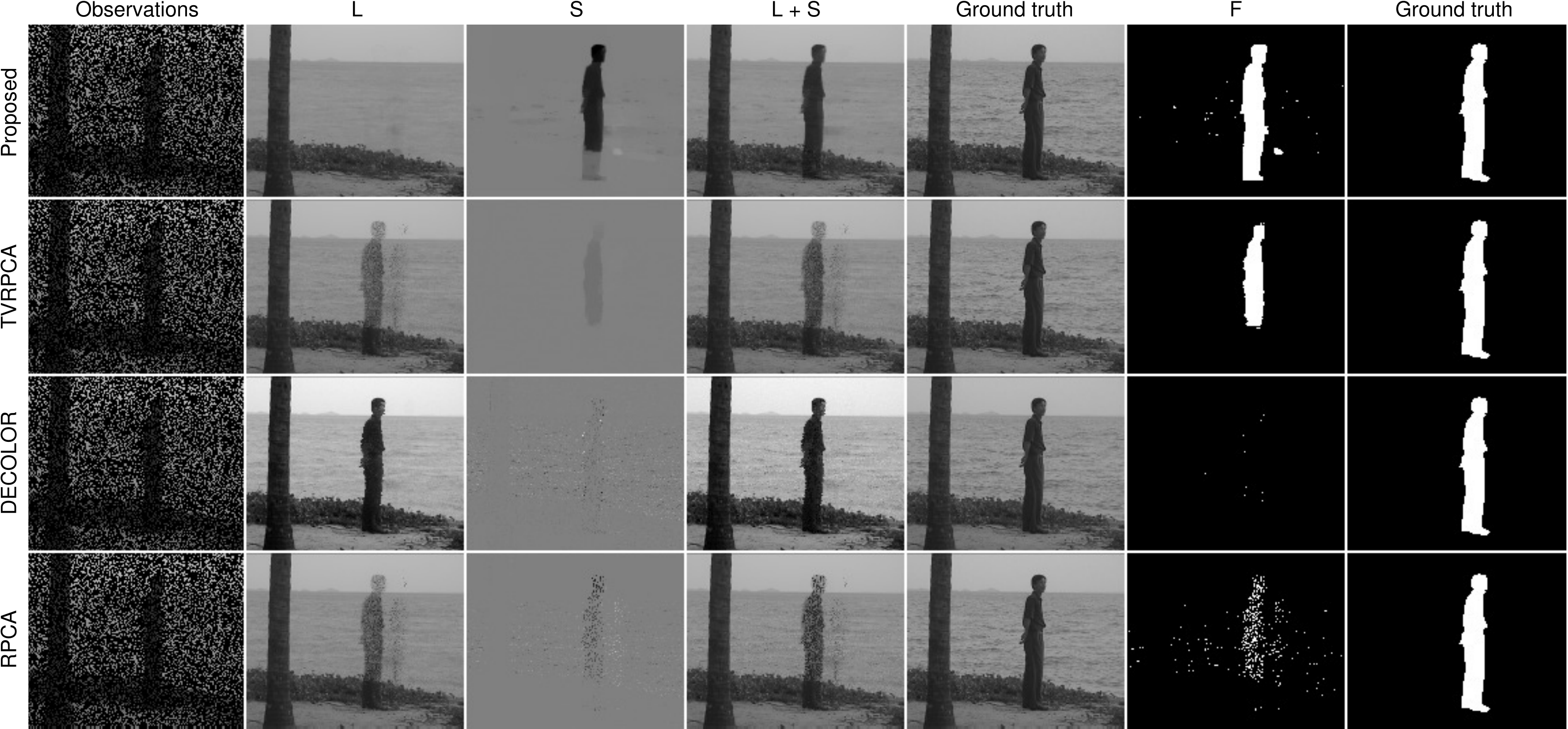}
\caption{A representative frame from the decompositions produced by each method applied to the Water Surface sequence with 70\% missing data. Left column: observations; $L$: reconstructed background; $S$: reconstructed foreground; $L+S$: reconstructed scene; fifth column: Water Surface sequence; $F$: estimated foreground mask; right column: true mask.}
\label{fig:water:missing}
\end{figure*}

Tables~\ref{tab:hall:outliers}, \ref{tab:hall:gaussian} and \ref{tab:hall:missing} show the performance of each method on the Hall sequence (static camera) as a function of outlier probability, noise SNR, and missing data probability, respectively. Our proposed PRPCA method clearly outperforms the existing methods across the range of corruption levels. Figure~\ref{fig:water:missing} illustrates the decompositions produced by each method on the Water Surface sequence corrupted by 70\% missing data. The foreground estimates produced by RPCA and DECOLOR are not able to impute the missing foreground pixels because their models lack a spatial continuity constraint. The TVRPCA method produces a more accurate foreground component, but its estimated background component contains some foreground artifacts that are not present in the proposed PRPCA method. These results show that our proposed method is better able to uncover the true foreground and background components of corrupted video.

\begin{figure*}[t!]
\centering\includegraphics[width=\textwidth]{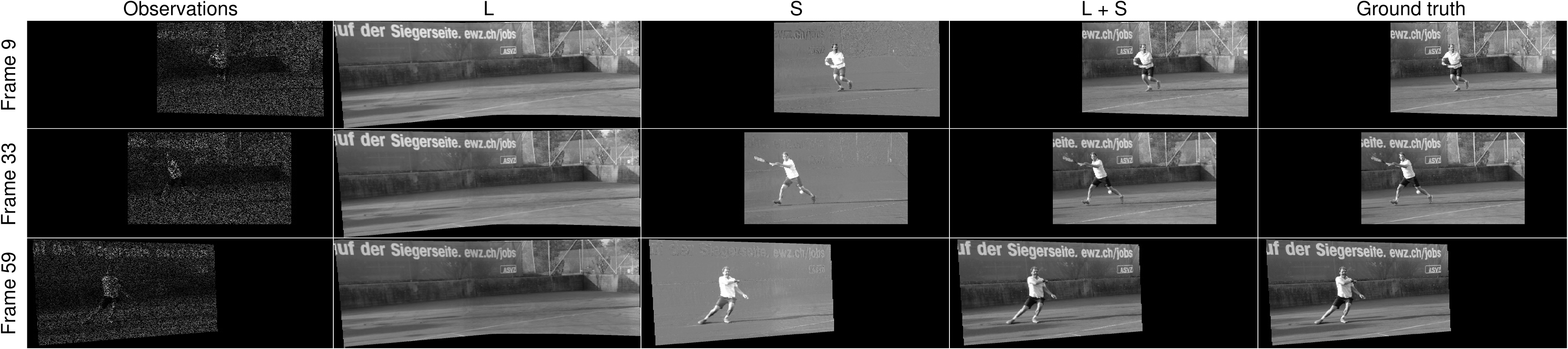}
\caption{Three representative frames from the decomposition produced by the proposed PRPCA method applied to the Tennis sequence corrupted by 70\% missing data. Left column: registered observations; $L$: reconstructed registered background; $S$: reconstructed registered foreground; $L+S$: reconstructed registered scene restricted to the current field of view; right column: registered Tennis sequence.}
\label{fig:tennis:missing:reg}
\end{figure*}

\begin{figure*}[t!]
\centering\includegraphics[width=\textwidth]{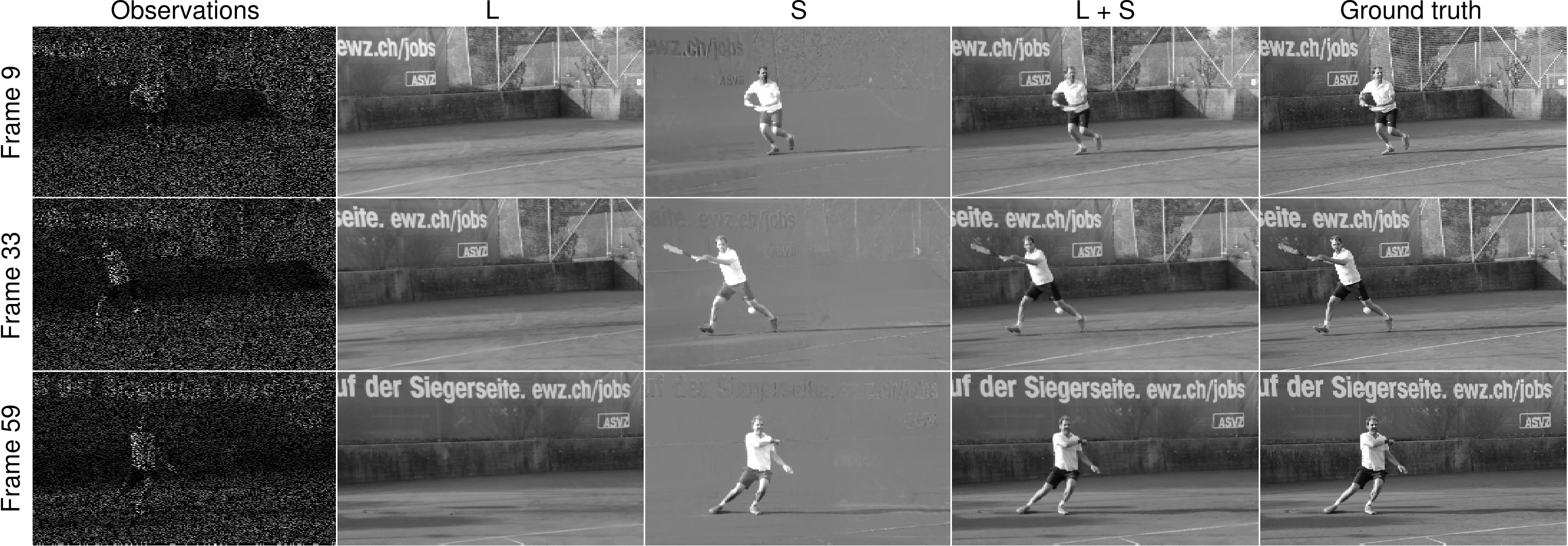}
\caption{The decompositions from Figure~\ref{fig:tennis:missing:reg} mapped to the perspective of the original video.}
\label{fig:tennis:missing}
\end{figure*}

Table~\ref{tab:davis:missing} shows the performance of each method on sequences from the DAVIS dataset corrupted by 70\% missing data. Tables~\ref{tab:tennis:outliers}, \ref{tab:tennis:poisson} and~\ref{tab:tennis:missing} show the performance of each method on the Tennis sequence as a function of outlier probability, SNR, and missing data probability, respectively. Our proposed PRPCA method clearly outperforms DECOLOR and achieves higher reconstruction PSNR than the baseline denoising methods. Figures~\ref{fig:tennis:missing:reg} depicts the decompositions produced by our proposed PRPCA method on the Tennis sequence corrupted by 70\% missing data. Note how our proposed method gracefully combines the background information from the corrupted frames to produce a clean panoramic estimate ($L$) of the full field of view. Also, the registered TV-regularized component ($S$) is able to accurately estimate the dynamic foreground from partial observations by exploiting its spatial continuity. Figure~\ref{fig:tennis:missing} shows the decompositions from Figure~\ref{fig:tennis:missing:reg} mapped to the perspective of the original video by applying the inverse homographies computed during frame registration. These sequences constitute a direct decomposition of the original moving camera video. In Figure~\ref{fig:tennis:missing}, the outline of the background text is faintly visible in Frames 9 and 33 of $S$. These artifacts arise from small mismatches in the frame registration process due to violations of the underlying far-field assumption of the frame registration model. This parallax effect captured by $S$ arises so that the reconstructed scene $L+S$ remains faithful to the data $Y$.

\begin{table*}[t!]
\resizebox{\textwidth}{!}{
\begin{tabular}{|c|ccc|ccc|ccc|ccc|}
\hline
\multirow{2}{*}{Sequence} &
\multicolumn{3}{c|}{Proposed} &
\multicolumn{3}{c|}{RPCA} &
\multicolumn{3}{c|}{TVRPCA} &
\multicolumn{3}{c|}{DECOLOR} \\
\cline{2-13}
& f-PSNR & b-PSNR & F-measure & f-PSNR & b-PSNR & F-measure & f-PSNR & b-PSNR & F-measure & f-PSNR & b-PSNR & F-measure \\
\hline \hline
Hall          & \textbf{37.25} & \textbf{36.58} &         0.58  & 27.64 & 30.75 & 0.27 & 31.02 & 32.58 & 0.35 & 29.69 & 33.17 & \textbf{0.65} \\
Fountain      & \textbf{37.78} & \textbf{34.52} & \textbf{0.70} & 29.59 & 26.90 & 0.24 & 36.04 & 29.62 & 0.32 & 32.51 & 26.23 &         0.56  \\
Escalator     & \textbf{30.87} & \textbf{28.95} & \textbf{0.70} & 21.85 & 23.05 & 0.30 & 24.09 & 24.89 & 0.38 & 23.53 & 24.99 &         0.41  \\
Water Surface & \textbf{40.00} & \textbf{34.99} & \textbf{0.93} & 31.93 & 29.50 & 0.33 & 33.57 & 30.03 & 0.70 & 28.79 & 18.42 &         0.17  \\
Shopping Mall & \textbf{37.70} & \textbf{39.87} &         0.73  & 28.03 & 32.62 & 0.35 & 31.70 & 34.09 & 0.46 & 29.65 & 34.05 & \textbf{0.76} \\
\hline \hline
Average       & \textbf{36.72} & \textbf{34.98} & \textbf{0.73} & 27.81 & 28.56 & 0.30 & 31.28 & 30.24 & 0.44 & 28.83 & 27.37 &         0.51  \\
\hline
\end{tabular}}
\caption{Performance metrics for each method on sequences from the I2R dataset corrupted by 70\% missing data.}
\label{tab:i2r:missing}
\end{table*}

\begin{table*}[t!]
\resizebox{\textwidth}{!}{
\begin{tabular}{|c|ccc|ccc|ccc|ccc|}
\hline
\multirow{2}{*}{p} &
\multicolumn{3}{c|}{Proposed} &
\multicolumn{3}{c|}{RPCA} &
\multicolumn{3}{c|}{TVRPCA} &
\multicolumn{3}{c|}{DECOLOR} \\
\cline{2-13}
& f-PSNR & b-PSNR & F-measure & f-PSNR & b-PSNR & F-measure & f-PSNR & b-PSNR & F-measure & f-PSNR & b-PSNR & F-measure \\
\hline
10\% & \textbf{41.48} & \textbf{39.37} & \textbf{0.60} & 30.35 & 32.67 & 0.27 & 38.38 & 38.98 & \textbf{0.60} & 30.28 & 31.54 & 0.29 \\
20\% & \textbf{38.94} & \textbf{37.98} & \textbf{0.60} & 27.12 & 32.63 & 0.19 & 36.50 & 37.42 & \textbf{0.60} & 27.02 & 31.63 & 0.17 \\
30\% & \textbf{37.69} & \textbf{36.21} & \textbf{0.59} & 25.40 & 32.39 & 0.15 & 34.94 & 36.08 &         0.58  & 30.27 & 31.54 & 0.29 \\
40\% & \textbf{36.49} & \textbf{34.73} & \textbf{0.58} & 24.26 & 32.03 & 0.13 & 32.51 & 24.13 &         0.57  & 24.13 & 18.50 & 0.07 \\
50\% & \textbf{35.84} & \textbf{33.73} & \textbf{0.57} & 23.57 & 31.49 & 0.12 & 29.85 & 18.11 &         0.49  & 23.47 & 14.61 & 0.07 \\
60\% & \textbf{34.93} & \textbf{32.38} & \textbf{0.56} & 22.87 & 31.36 & 0.10 & 27.98 & 14.65 &         0.35  & 22.79 & 14.13 & 0.07 \\
\hline
\end{tabular}}
\caption{Performance metrics for each method on the Hall sequence as a function of outlier probability.}
\label{tab:hall:outliers}
\end{table*}

\begin{table*}[t!]
\resizebox{\textwidth}{!}{
\begin{tabular}{|c|ccc|ccc|ccc|ccc|}
\hline
\multirow{2}{*}{SNR} &
\multicolumn{3}{c|}{Proposed} &
\multicolumn{3}{c|}{RPCA} &
\multicolumn{3}{c|}{TVRPCA} &
\multicolumn{3}{c|}{DECOLOR} \\
\cline{2-13}
& f-PSNR & b-PSNR & F-measure & f-PSNR & b-PSNR & F-measure & f-PSNR & b-PSNR & F-measure & f-PSNR & b-PSNR & F-measure \\
\hline
5 dB  & \textbf{31.78} & \textbf{26.15} & \textbf{0.52} & 20.85 & 18.55 & 0.07 & 25.20 & 11.29 &         0.08  & 27.98 & 14.30 &         0.07  \\
10 dB & \textbf{32.78} & \textbf{27.87} & \textbf{0.54} & 23.04 & 23.31 & 0.08 & 26.85 & 13.33 &         0.14  & 28.54 & 14.30 &         0.07  \\
20 dB & \textbf{34.73} & \textbf{30.73} & \textbf{0.56} & 27.42 & 28.73 & 0.14 & 30.20 & 16.89 &         0.34  & 30.13 & 14.30 &         0.07  \\
30 dB & \textbf{36.66} & \textbf{32.72} &         0.58  & 31.80 & 30.14 & 0.30 & 34.64 & 21.83 & \textbf{0.59} & 31.65 & 25.14 &         0.56  \\
40 dB & \textbf{39.64} & \textbf{33.90} & \textbf{0.60} & 36.20 & 31.27 & 0.46 & 37.96 & 25.70 &         0.58  & 36.27 & 31.51 &         0.59  \\
50 dB & \textbf{42.89} & \textbf{36.14} &         0.60  & 40.59 & 32.00 & 0.54 & 41.47 & 29.77 &         0.59  & 37.87 & 32.73 & \textbf{0.61} \\
\hline
\end{tabular}}
\caption{Performance metrics for each method on the Hall sequence as a function of SNR (Gaussian noise).}
\label{tab:hall:gaussian}
\end{table*}

\begin{table*}[t!]
\resizebox{\textwidth}{!}{
\begin{tabular}{|c|ccc|ccc|ccc|ccc|}
\hline
\multirow{2}{*}{p} &
\multicolumn{3}{c|}{Proposed} &
\multicolumn{3}{c|}{RPCA} &
\multicolumn{3}{c|}{TVRPCA} &
\multicolumn{3}{c|}{DECOLOR} \\
\cline{2-13}
& f-PSNR & b-PSNR & F-measure & f-PSNR & b-PSNR & F-measure  & f-PSNR & b-PSNR & F-measure & f-PSNR & b-PSNR & F-measure \\
\hline
60\% & \textbf{39.01} & \textbf{37.79} &         0.59  & 28.33 & 31.19 & 0.33 & 35.44 & 36.01 & 0.50 & 30.35 & 32.57 & 0.64 \\
70\% & \textbf{37.25} & \textbf{36.58} &         0.58  & 27.64 & 30.75 & 0.27 & 31.02 & 32.58 & 0.35 & 29.69 & 33.17 & 0.65 \\
80\% & \textbf{35.69} & \textbf{35.43} &         0.58  & 27.13 & 30.00 & 0.20 & 30.32 & 33.06 & 0.07 & 28.26 & 31.47 & 0.23 \\
90\% & \textbf{33.30} & \textbf{33.40} & \textbf{0.55} & 27.45 & 23.43 & 0.08 & 30.26 & 14.50 & 0.07 & 27.13 & 31.19 & 0.11 \\
\hline
\end{tabular}}
\caption{Performance metrics for each method on the Hall sequence as a function of missing data probability.}
\label{tab:hall:missing}
\end{table*}

\begin{table*}[t!]
\centering\resizebox{0.75\textwidth}{!}{
\begin{tabular}{|c|ccc|ccc|cc|}
\hline
\multirow{2}{*}{Sequence} &
\multicolumn{3}{c|}{Proposed} &
\multicolumn{3}{c|}{DECOLOR} &
\multicolumn{2}{c|}{Baseline (Interpolation)} \\
\cline{2-9}
& f-PSNR & b-PSNR & F-measure & f-PSNR & b-PSNR & F-measure & f-PSNR & b-PSNR \\
\hline \hline
Tennis      & \textbf{40.50} & \textbf{30.86} & \textbf{0.77} & 22.38 & 17.76 &         0.37  & 40.02 & 29.86 \\
Paragliding & \textbf{43.33} & \textbf{34.59} &         0.77  & 27.13 & 18.86 & \textbf{0.83} & 42.84 & 32.95 \\
Rollerblade & \textbf{42.48} & \textbf{29.65} & \textbf{0.83} & 24.98 & 19.68 &         0.78  & 41.90 & 27.87 \\
Horsejump   & \textbf{36.49} & \textbf{27.70} & \textbf{0.76} & 23.28 & 17.83 &         0.29  & 36.19 & 25.93 \\
\hline \hline
Average     & \textbf{40.70} & \textbf{30.70} & \textbf{0.78} & 24.44 & 18.53 &         0.57  & 40.24 & 29.15 \\
\hline
\end{tabular}}
\caption{Performance metrics for each method on sequences from the DAVIS dataset corrupted by 70\% missing data.}
\label{tab:davis:missing}
\end{table*}

\begin{table*}[t!]
\centering\resizebox{0.65\textwidth}{!}{
\begin{tabular}{|c|ccc|ccc|cc|}
\hline
\multirow{2}{*}{p} &
\multicolumn{3}{c|}{Proposed} &
\multicolumn{3}{c|}{DECOLOR} &
\multicolumn{2}{c|}{Baseline (Median Filter)} \\
\cline{2-9}
& f-PSNR & b-PSNR & F-measure & f-PSNR & b-PSNR & F-measure & f-PSNR & b-PSNR \\
\hline
10\% & \textbf{41.88} & \textbf{30.98} & \textbf{0.76} & - & - & - & 41.44 & 30.54 \\
20\% & \textbf{41.04} & \textbf{30.46} & \textbf{0.76} & - & - & - & 40.55 & 29.61 \\
30\% & \textbf{39.39} & \textbf{30.13} & \textbf{0.76} & - & - & - & 38.58 & 27.36 \\
40\% & \textbf{38.21} & \textbf{29.82} & \textbf{0.7}5 & - & - & - & 35.76 & 24.21 \\
50\% & \textbf{36.86} & \textbf{29.14} & \textbf{0.73} & - & - & - & 32.72 & 21.04 \\
\hline
\end{tabular}}
\caption{Performance metrics for each method on the Tennis sequence as a function of outlier probability. DECOLOR raises an error due to the significant camera motion, so it produces no decompositions.}
\label{tab:tennis:outliers}
\end{table*}

\begin{table*}[t!]
\centering\resizebox{0.7\textwidth}{!}{
\begin{tabular}{|c|ccc|ccc|cc|}
\hline
\multirow{2}{*}{p} &
\multicolumn{3}{c|}{Proposed} &
\multicolumn{3}{c|}{DECOLOR} &
\multicolumn{2}{c|}{Baseline (Interpolation)} \\
\cline{2-9}
& f-PSNR & b-PSNR & F-measure & f-PSNR & b-PSNR & F-measure & f-PSNR & b-PSNR \\
\hline
60\% & \textbf{41.61} & \textbf{31.33} & \textbf{0.77} & 22.09 & 17.35 & 0.36 & 41.23 & 30.60 \\
70\% & \textbf{40.50} & \textbf{30.86} & \textbf{0.77} & 22.38 & 17.76 & 0.37 & 40.02 & 29.86 \\
80\% & \textbf{38.74} & \textbf{30.28} & \textbf{0.75} & 22.48 & 17.47 & 0.43 & 38.35 & 28.84 \\
90\% & \textbf{35.78} & \textbf{29.22} & \textbf{0.74} & 22.95 & 17.54 & 0.38 & 35.67 & 27.18 \\
\hline
\end{tabular}}
\caption{Performance metrics for each method on the Tennis sequence as a function of missing data probability.}
\label{tab:tennis:missing}
\end{table*}

\section{OptShrink background} \label{app:optshrink:background}
Here we provide some additional detail about the OptShrink estimator that we introduced in Section~\ref{subsec:optshrink}. We begin by motivating the need for OptShrink by discussing the suboptimality of singular value thresholding (SVT) for low-rank matrix denoising, and then we explicitly describe the estimator.

One can view the SVT-based low-rank update
\begin{equation} \label{Lupdate2}
L^{k+1} = \SVT_{\tau^k \lambda_L}(L^k - \tau^{k} U^{k+1})
\end{equation}
from \eqref{eq:prox:updates:2} as a low-rank denoising step, where the matrix $L^k - \tau^{k} U^{k+1}$ is a noisy version of a latent low-rank matrix $L_{\text{true}}$ that we are interested in recovering, and the SVT is the chosen low-rank estimator.

A natural question to ask is what is the quality of the low-rank estimates produced by the SVT operator. To address this question, suppose that we are given a matrix $\widetilde{X} \in \mathbb{R}^{m \times n}$ of the form
\begin{equation} \label{eq:lowrankmodel2}
\widetilde{X} = \underbrace{\sum_{i=1}^{r} \theta_i u_i v_i ^{H}}_{=:L} + X,
\end{equation}
where $L$ is an unknown rank-$r$ matrix with singular values $\theta_i$ and singular vectors $u_i$ and $v_i$, and $X$ is an additive noise matrix. For example, in \eqref{Lupdate2}, we identity $L = L_{\text{true}}$ and $X = L^k - L_{\text{true}} - \tau^{k} U^{k+1}$.

Now, consider the oracle low-rank denoising problem
\begin{equation} \label{eq:denoising2}
w^{\star} = \arg \min_{[w_{1}, \ldots, w_{r}]^{T} \in \mathbb{R}^{r}} \Big\| \sum_{i=1}^{r} \theta_i u_i v_i ^{H} - \sum_{i=1}^{r} w_{i} \widetilde{u}_{i} \widetilde{v}_{i}^{H} \Big\|_{F},
\end{equation}
where $\widetilde{u}_i$ and $\widetilde{v}_i$ are the singular vectors of $\widetilde{X}$, and we denote its singular values by $\widetilde{\sigma}_i$. Problem \eqref{eq:denoising2} seeks the best approximation of the latent low-rank signal matrix $L$ by an optimally weighted combination of estimates of its left and right singular vectors. The truncated SVD (of rank $r$) and SVT are both feasible points for \eqref{eq:denoising2}. Indeed,  the truncated SVD corresponds to choosing weights $w_i = \widetilde{\sigma}_i \mathds{1}\{i \leq r\}$ and SVT with parameter $\tau \geq \widetilde{\sigma}_{r+1}$ corresponds to $w_i = (\widetilde{\sigma}_i - \tau)^+$. However, \eqref{eq:denoising2} can be solved in closed-form (see \cite{nadakuditi2014optshrink}), yielding the expression
\begin{equation}\label{eq:denoising solution2}
w^{\star}_{i} = \sum_{j=1}^{r} \theta_j\left(\widetilde{u}_{i}^{H} u_{j}\right) \left(\widetilde{v}_{i}^H v_{j}\right), \quad i=1,\ldots,r.
\end{equation}
Of course, \eqref{eq:denoising solution2} cannot be computed in practice because it depends on the latent low-rank singular vectors $u_i$ and $v_i$ that we would like to estimate, but it gives insight into the properties of the optimal weights $w^{\star}$. Indeed, when $\widetilde{u}_{i}$ and $\widetilde{v}_{i}$ are good estimates of $u_{i}$ and $v_i$, respectively, we expect $\widetilde{u}_{i}^{H} u_{i}$ and $\widetilde{v}_{i}^H v_{i}$ to be close to $1$. Consequently, from \eqref{eq:denoising solution2}, we expect ${w}^{\star}_{i}  \approx \theta_{i}$. Conversely, when $\widetilde{u}_{i}$ and $\widetilde{v}_{i}$ are poor estimates of $u_{i}$ and $v_{i}$, respectively, we expect $\widetilde{u}_{i}^{H} u_{i}$ and $v_{i}^{H} \widetilde{v}_{i}$ to be closer to $0$ and ${w}^{\star}_{i} < \theta_i$. In other words, \eqref{eq:denoising solution2} shows that the optimal singular value shrinkage is inversely proportional to the accuracy of the estimated principal subspaces. As a special case, if $\theta_i \rightarrow \infty$, then clearly $\widetilde{u}_{i}^{H} u_{i} \to 1$ and $v_{i}^{H} \widetilde{v}_{i} \to 1$, so the optimal weights $w^{\star}_i$ must have the property that the absolute shrinkage vanishes as $\theta_{i} \to \infty$. Consequently, the SVT operator, which applies a constant shrinkage to each singular value of its input, will necessarily produce suboptimal low-rank estimates in general. See \cite{nadakuditi2014optshrink} for more details.

The following theorem \cite{nadakuditi2014optshrink} formalizes the above argument under a probabilistic model for the additive noise matrix $X$.

\begin{theorem} \label{th:optshrink}
Suppose that $X_{ij}$ are i.i.d. random variables with zero-mean, variance $\sigma^2$, and bounded higher order moments, and suppose that $\theta_1 > \theta_2 > \ldots > \theta_r > \sigma$. Then, as $m,n \to \infty$ such that $m/n \to c \in (0, \infty)$, we have that
\begin{equation} \label{eq:asympweights}
{w}^{\star}_{i}  + 2 \ \displaystyle\frac{D_{\mu_{\widetilde{X}}}(\widetilde{\sigma}_i)}{D'_{\mu_{\widetilde{X}}}(\widetilde{\sigma}_i)} \convas 0 \qquad \textrm{for} \ \ i = 1, \ldots, r,
\end{equation}
where
\begin{equation}
\mu_{\widetilde{X}}(t) = \frac{1}{q-r} \sum_{i=r+1}^{q} \delta\left(t - \widetilde{\sigma}_i\right),
\end{equation}
with $q = \min(m,n)$ and the $D$-transform is defined as
\begin{equation} \label{eq:Dtransform2}
\begin{array}{rl}
D_{\mu_{\widetilde{X}}}(z) := &\left[ \displaystyle\int \frac{z}{z^2 - t^2} \ \mathrm{d}\mu_{\widetilde{X}}(t) \right] \times \\ \vphantom{2^{2^{2^{2^{2^{2^2}}}}}} &\left[ c \displaystyle\int \frac{z}{z^2 - t^2} \ \mathrm{d}\mu_{\widetilde{X}}(t) + \frac{1 - c}{z} \right],
\end{array}
\end{equation}
and $D_{\mu_{\widetilde{X}}}'(z)$ is the derivative of $D_{\mu_{\widetilde{X}}}(z)$ with respect to $z$.
\end{theorem}

Theorem~\ref{th:optshrink} establishes that the optimal weights ${w}^{\star}_{i}$ converge in the large matrix limit to a certain non-random integral transformation of the limiting noise distribution $\mu_{\widetilde{X}}$.

In practice, Theorem~\ref{th:optshrink} also suggests the following data-driven OptShrink estimator, defined for a given matrix $Y \in \mathbb{C}^{m \times n}$ and rank $r$ as
\begin{equation} \label{eq:optshrink_defn2}
\mathbf{OptShrink}_r(Y) = \sum_{i=1}^{r}  \left(-2 \ \displaystyle\frac{D_{\mu_Y}(\sigma_i)}{D'_{\mu_Y}(\sigma_i)} \right) u_i v_i^{H},
\end{equation}
where $Y = U  \Sigma V^H$ is the SVD of $Y$ with singular values $\sigma_i$, and
\begin{equation} \label{eq:empmassfcn2}
\mu_{Y}(t) = \frac{1}{q-r} \sum_{i=r+1}^{q} \delta\left(t - \sigma_i\right)
\end{equation}
is the empirical mass function of the noise-only singular values of $Y$ with $q = \min(m,n)$. By Theorem~\ref{th:optshrink}, $\mathbf{OptShrink}_r(\widetilde{X})$ asymptotically solves the oracle denoising problem \eqref{eq:denoising2}.

OptShrink has a single parameter $r \in \mathbb{N}$ that directly specifies the rank of its output matrix. Rather than applying a constant shrinkage to each singular value of the input matrix as in SVT, the OptShrink estimator partitions the singular values of its input matrix into \emph{signals} $\{\sigma_1,\ldots,\sigma_r\}$ and \emph{noise} $\{\sigma_{r+1},\ldots,\sigma_q\}$ and uses the empirical mass function of the noise singular values to estimate the optimal (nonlinear, in general) shrinkage \eqref{eq:asympweights} to apply to each signal singular value. See \cite{nadakuditi2014optshrink,benaych2012singular} for additional detail.

The computational cost of OptShrink is the cost of computing a full SVD\footnote{In practice, one need only compute the singular values $\sigma_1,\ldots,\sigma_q$ and the leading $r$ singular vectors of $Y$.} plus the $O(r(m+n))$ computations required to compute the $D$-transform terms in \eqref{eq:optshrink_defn2}, which reduce to summations for the choice of ${\mu}_{Y}$ in \eqref{eq:empmassfcn2}.

\section{Incorporating Missing Data} \label{app:missingdata}
In this section we describe how we adapt the RPCA, DECOLOR, and TVRPCA algorithms for our inpainting experiments in Appendix~\ref{app:additional:experiments}. Throughout, we use $Y \in \R^{mn \times p}$ to denote the matrix whose columns contain the vectorized frames of the input video with missing data.

\subsection{RPCA} \label{subsec:RPCA}
The standard robust PCA \cite{candes2011robust,otazo2015low} method minimizes the cost
\begin{equation} \label{eq:rpca}
\min_{L,S} ~ \tfrac{1}{2} \|Y - L - S\|_F^2 + \lambda_L \|L\|_{\star} + \lambda_S\|S\|_1,
\end{equation}
where $L$ is the low-rank background component and $S$ is the sparse foreground component. We incorporate a missing data mask into \eqref{eq:rpca} analogously to our approach in our proposed method; that is, we solve the modified RPCA problem
\begin{equation} \label{eq:rpca:missing}
\min_{L,S} ~ \tfrac{1}{2} \|\proj_{M}(Y - L - S)\|_F^2 + \lambda_L \|L\|_{\star} + \lambda_S\|S\|_1,
\end{equation}
where the missing data mask $M \in \{0,1\}^{mn \times d}$ with entries
\begin{equation} \label{eq:missing:mask}
M_{ij} = \begin{cases}
0 & Y_{ij} \text{ is missing} \\
1 & Y_{ij} \text{ is observed}
\end{cases}
\end{equation}
omits unobserved pixels from the data fidelity term in \eqref{eq:rpca}. Applying the same proximal gradient strategy to \eqref{eq:rpca:missing} as for the standard RPCA problem \eqref{eq:rpca} leads to the updates
\begin{equation} \label{eq:rpca:missing:updates}
\begin{array}{r@{~}c@{~}l}
Z^{k+1} &=& \proj_{M}(L^k + S^k - Y) \\[4pt]
L^{k+1} &=& \SVT_{\tau\lambda_L}(L^k - \tau Z^{k+1}) \\[4pt]
S^{k+1} &=& \soft_{\tau\lambda_S}(S^k - \tau Z^{k+1}),
\end{array}
\end{equation}
with constant step size $\tau^k = \tau < 1$ sufficing to guarantee convergence \cite{parikh2014proximal}. One can view the updates \eqref{eq:rpca:missing:updates} as a special case of our proposed updates when the camera is static (so that no frame registration is performed) and the TV regularization parameter tends to infinity.

\subsection{DECOLOR} \label{subsec:decolor}
The DECOLOR method minimizes the cost from Equation (20) of \cite{zhou2013moving}, which, in our notation, is
\begin{equation} \label{eq:decolor}
\begin{array}{rl}
\displaystyle\min_{\tau,L,S} & \tfrac{1}{2}\|\proj_{S^{\perp}}(Y \circ \tau - L)\|_F^2 ~+ \\
& \quad \quad \alpha\|L\|_{\star} + \beta\|S\|_1 + \gamma \TV(S),
\end{array}
\end{equation}
where $L$ is the low-rank (registered) background, $S_{ij} \in \{0,1\}$ is the (registered) foreground mask, $S^{\perp}$ is the orthogonal complement of $S$, $\tau$ are the 2D parametric transforms that register the input frames $Y$, and $\TV(\cdot)$ denotes unweighted anisotropic total variation.

The DECOLOR algorithm proceeds by alternating between updating $\tau$, $L$, and $S$ sequentially with all other variables held fixed. The $\tau$ subproblem is approximately solved using an iterative strategy where one linearizes \eqref{eq:decolor} with respect to $\tau$, solves the resulting weight least-squares problem, and then repeats the process to refine $\tau$. The $L$ subproblem for \eqref{eq:decolor} is a missing data version of the proximal operator for the nuclear norm and can be approximately solved by performing a few iterations of the SOFT-IMPUTE algorithm \cite{mazumder2010spectral}. Finally, the $S$ subproblem is a Markov random field problem that is solved exactly via graph cuts \cite{zhou2013moving}.

At any given step of the DECOLOR algorithm, the matrix $Y \circ \tau$ denotes the current estimate of the registered frames, so the appropriate missing data mask to consider is
\begin{equation}
M_{ij} = \begin{cases}
0 & \text{if } [Y \circ \tau]_{ij} \text{ is missing} \\
1 & \text{if } [Y \circ \tau]_{ij} \text{ is observed},
\end{cases}
\end{equation}
which implicitly depends on the current value of the parameteric transformations $\tau$. Thus, to incorporate this mask into \eqref{eq:decolor}, we solve the modified problem
\begin{equation} \label{eq:decolor:missing}
\begin{array}{rl}
\displaystyle\min_{\tau,L,S} & \tfrac{1}{2}\|\proj_{S^{\perp} \odot M}(Y \circ \tau - L))\|_F^2 ~+ \\
& \quad \quad \alpha\|L\|_{\star} + \beta\|S\|_1 + \gamma \TV(S),
\end{array}
\end{equation}
where $\odot$ denotes elementwise multiplication. Our modified problem \eqref{eq:decolor:missing} omits unobserved data in the registered perspective defined by $\tau$ from the data fidelity term. Note that we have the relation $\proj_{S^{\perp} \odot M}(\cdot) = \proj_{S^{\perp}}(\proj_{M}(\cdot)) = \proj_{M}(\proj_{S^{\perp}}(\cdot))$, which can be used to appropriately isolate $S$ in the projection operators when minimizing \eqref{eq:decolor:missing} with respect to $S$.

The same alternating minimization algorithm proposed in Algorithm~1 of \cite{zhou2013moving} can be extended to solve \eqref{eq:decolor:missing}. Indeed, after linearizing \eqref{eq:decolor:missing} around $\tau$, the inner iterations for updating $\tau$ can be written as
\begin{equation} \label{eq:decolor:missing:tau}
\tau^{k+1} = \tau^k + \arg \min_{\Delta \tau} ~ \|\proj_{S^{\perp} \odot M}(Y \circ \tau - L + J_{\tau^k} \Delta \tau)\|_F^2,
\end{equation}
where $J_{\tau}$ denotes the Jacobian matrix of \eqref{eq:decolor:missing} with respect to $\tau$. The iteration \eqref{eq:decolor:missing:tau} is still a weighted least squares problem that can be solved in closed-form. The $L$ subproblem can be approximately solved by performing a few inner iterations of the SOFT-IMPUTE updates
\begin{equation} \label{eq:decolor:missing:L}
L^{k+1} = \SVT_{\alpha}\left(\proj_{S^{\perp} \odot M}(Y \circ \tau) + \proj_{(S^{\perp} \odot M)^{\perp}}(L^k)\right).
\end{equation}
Finally, the $S$ subproblem can be written as
\begin{equation} \label{eq:decolor:missing:S}
\displaystyle\min_{S} \displaystyle\sum_{ij} \left(\beta - \tfrac{1}{2}[\proj_{M}(Y \circ \tau - L)]_{ij}^2\right)S_{ij} + \gamma \TV(S),
\end{equation}
which can be solved using the same graph cuts algorithm from \cite{zhou2013moving} with residual matrix $\proj_{M}(Y \circ \tau - L)$ in place of $Y \circ \tau - L$.

Aside from the modified subproblem updates in \eqref{eq:decolor:missing:tau}-\eqref{eq:decolor:missing:S}, we retain all other features of the DECOLOR method as outlined in Algorithm~1 of \cite{zhou2013moving}. Note that the above updates reduce to the original DECOLOR algorithm when $M$ is the all-ones matrix (no missing data).

\subsection{TVRPCA} \label{subsec:tvrpca}
The TVRPCA method minimizes the cost from Equation (7) of \cite{cao2016total}, which, in our notation, is
\begin{equation} \label{eq:tvrpca}
\begin{array}{rl}
\displaystyle\min_{L,G,E,S} & ~ \|L\|_{\star} + \lambda_1\|G\|_1 + \lambda_2\|E\|_1 + \lambda_3 \TV(S) \\[8pt]
\text{s.t.} & ~ Y = L + G, ~ G = E + S.
\end{array}
\end{equation}
In \eqref{eq:tvrpca}, $L$ is the low-rank background component and $G$ is a residual component, which is further decomposed into a smooth foreground component $S$ and a sparse error term $E$. The authors propose to solve \eqref{eq:tvrpca} by applying an alternating minimization scheme to the augmented Lagrangian of \eqref{eq:tvrpca}:
\begin{equation} \label{eq:tvrpca:lagrangian}
\begin{multlined}
\mathcal{L}_{\mu}(L,G,E,S,X,Z) ~= \\
\|L\|_{\star} + \lambda_1\|G\|_1 + \lambda_2\|E\|_1 + \lambda_3 \TV(S) ~+ \\
\tfrac{\mu}{2}\|Y - L - G\|_F^2 + \langle X,~Y-L-G \rangle ~+ \\
\tfrac{\mu}{2}\|G - E - S\|_F^2 + \langle Z,~G-E-S \rangle.
\end{multlined}
\end{equation}
In particular, in \cite{cao2016total} one sequentially updates each component $\{L,G,E,S,X,Z\}$ by minimizing \eqref{eq:tvrpca:lagrangian} with all other components held fixed.

We incorporate a missing data mask into \eqref{eq:tvrpca} by solving the related problem
\begin{align} \label{eq:tvrpca:missing}
\min_{L,G,E,S} & ~ ~ \|L\|_{\star} + \lambda_1\|G\|_1 + \lambda_2\|E\|_1 + \lambda_3 \TV(S) \nonumber \\
\text{s.t.} & ~ ~ \proj_{M}(Y) = \proj_{M}(L + G), ~ G = E + S,
\end{align}
which omits equality constraints involving unobserved pixels from \eqref{eq:tvrpca:missing}. The augmented Lagrangian for \eqref{eq:tvrpca:missing} is
\begin{equation} \label{eq:tvrpca:missing:lagrangian}
\begin{multlined}
\mathcal{L}_{\mu}(L,G,E,S,X,Z) ~= \\
\|L\|_{\star} + \lambda_1\|G\|_1 + \lambda_2\|E\|_1 + \lambda_3 \TV(S) ~+ \\
\tfrac{\mu}{2}\|\proj_{M}(Y - L - G)\|_F^2 + \langle X, ~\proj_{M}(Y-L-G) \rangle ~+ \\
\tfrac{\mu}{2}\|G - E - S\|_F^2 + \langle Z, ~G-E-S \rangle,
\end{multlined}
\end{equation}
and we solve \eqref{eq:tvrpca:missing} by applying the same alternating minimization strategy to \eqref{eq:tvrpca:missing:lagrangian} from the TVRPCA method. The subproblem updates for minimizing \eqref{eq:tvrpca:missing:lagrangian} are the same as those derived in Section~III-C of \cite{cao2016total} for the original cost \eqref{eq:tvrpca}, with the following modifications.\footnote{In the modified $L$ and $G$ updates, we assume that the initial $X^0$ satisfies $\proj_{M^{\perp}}(X^0) = 0$, which is true when one chooses $X^0 = 0$.} Fist, the $L$ subproblem for \eqref{eq:tvrpca:missing:lagrangian} can be written in the form of a SOFT-IMPUTE problem \cite{mazumder2010spectral}, so it can be approximately solved using a few inner iterations of the updates
\begin{equation} \label{eq:tvrpca:missing:L}
L^{k+1} = \SVT_{\frac{1}{\mu}}(\proj_{M}(Y-G+\tfrac{1}{\mu}X) + \proj_{M^{\perp}}(L^k)).
\end{equation}
After suitable manipulation, the $G$ subproblem for \eqref{eq:tvrpca:missing:lagrangian} can be written as two disjoint soft-thresholding problems with different shrinkage parameters. Indeed, the minimizer $\hat{G}$ of \eqref{eq:tvrpca:missing:lagrangian} with respect to $G$ can be written as
\begin{align} \label{eq:tvrpca:missing:G}
&\proj_{M}(\hat{G}) = \proj_{M}\bigg[\soft_{\frac{\lambda_1}{2\mu}}\Big(\tfrac{1}{2}(Y - L + E + S) + \tfrac{1}{2\mu} (X - Z))\Big)\bigg] \nonumber \\
&\proj_{M^{\perp}}(\hat{G}) = \proj_{M^{\perp}}\left[\soft_{\frac{\lambda_1}{\mu}}\left(E + S - \tfrac{1}{\mu}Z\right)\right].
\end{align}
Finally, the $X$ subproblem for \eqref{eq:tvrpca:missing:lagrangian} can be solved exactly using the simple update
\begin{equation} \label{eq:tvrpca:missing:X}
X \leftarrow X + \mu \proj_{M}(Y - L - G).
\end{equation}
All other subproblems for \eqref{eq:tvrpca:missing} are identical to the method outlined in Section III-C of \cite{cao2016total} for the original cost \eqref{eq:tvrpca}. Note that the above updates reduce to the original TVRPCA algorithm when $M$ is the all-ones matrix (no missing data).

\bibliographystyle{IEEEtran}
\bibliography{references}

\end{document}